\documentclass[journal]{IEEEtran}
\usepackage{graphicx}
\usepackage{cite}
\usepackage{epstopdf}
\usepackage{amsfonts}
\usepackage{amsmath}
\usepackage{algorithm}
\usepackage{algorithmic}
\usepackage{subfig}
\usepackage{array}
\usepackage{multicol}
\usepackage{multirow}
\usepackage{enumerate}
\usepackage{epsfig}
\usepackage{amssymb}
\usepackage{bbding}

\begin{document}

\title{TCGL: Temporal Contrastive Graph for Self-supervised Video Representation Learning}

\author{Yang~Liu,~\IEEEmembership{Member,~IEEE},
        Keze~Wang,
        Lingbo~Liu,
        Haoyuan~Lan,
        and~Liang~Lin,~\IEEEmembership{Senior~Member,~IEEE}
\thanks{This work is supported in part by the National Natural Science Foundation of China under Grant No.62002395, in part by the National Natural Science Foundation of Guangdong Province (China) under Grant No. 2021A15150123, and in part by the China Postdoctoral Science Foundation funded project under Grant No.2020M672966. (\emph{Corresponding author: Keze Wang.}) }
\thanks{Yang Liu, Keze Wang, Haoyuan Lan and Liang Lin are with the School
of Computer Science and Engineering, Sun Yat-sen University, Guangzhou, China. \protect
(E-mail: liuy856@mail.sysu.edu.cn, kezewang@gmail.com, lanhy5@mail2.sysu.edu.cn, linliang@ieee.org)

Lingbo Liu is with The Hong Kong Polytechnic University, HongKong, China. (Email: lingbo.liu@polyu.edu.hk)}}

\markboth{ IEEE TRANSACTIONS ON IMAGE PROCESSING}%
{Shell \MakeLowercase{\textit{et al.}}: Bare Demo of IEEEtran.cls for IEEE Journals}
\maketitle

\begin{abstract}
Video self-supervised learning is a challenging task, which requires significant expressive power from the model to leverage rich spatial-temporal knowledge and generate effective supervisory signals from large amounts of unlabeled videos. However, existing methods fail to increase the temporal diversity of unlabeled videos and ignore elaborately modeling multi-scale temporal dependencies in an explicit way. To overcome these limitations, we take advantage of the multi-scale temporal dependencies within videos and proposes a novel video self-supervised learning framework named Temporal Contrastive Graph Learning (TCGL), which jointly models the inter-snippet and intra-snippet temporal dependencies for temporal representation learning with a hybrid graph contrastive learning strategy. Specifically, a Spatial-Temporal Knowledge Discovering (STKD) module is first introduced to extract motion-enhanced spatial-temporal representations from videos based on the frequency domain analysis of discrete cosine transform. To explicitly model multi-scale temporal dependencies of unlabeled videos, our TCGL integrates the prior knowledge about the frame and snippet orders into graph structures, i.e., the intra-/inter- snippet Temporal Contrastive Graphs (TCG). Then, specific contrastive learning modules are designed to maximize the agreement between nodes in different graph views. To generate supervisory signals for unlabeled videos, we introduce an Adaptive Snippet Order Prediction (ASOP) module which leverages the relational knowledge among video snippets to learn the global context representation and recalibrate the channel-wise features adaptively. Experimental results demonstrate the superiority of our TCGL over the state-of-the-art methods on large-scale action recognition and video retrieval benchmarks. The code is publicly available at https://github.com/YangLiu9208/TCGL.
\end{abstract}


\maketitle

\IEEEpeerreviewmaketitle

\section{Introduction}\label{sec:introduction}
\IEEEPARstart{D}{eep} convolutional neural networks (CNNs) \cite{krizhevsky2012imagenet} have achieved state-of-the-art performance in many visual recognition tasks. This can be primarily attributed to the learned rich representation from well-trained networks using large-scale image/video datasets (e.g. ImageNet \cite{deng2009imagenet}, Kinetics \cite{kay2017kinetics}, SomethingSomething \cite{goyal2017something}) with strong supervision information \cite{kim2019self}. However, annotating such large-scale data is laborious, expensive, and impractical, especially for complex data-based high-level tasks, such as video action understanding and retrieval. To fully leverage the large amount of unlabeled data, self-supervised learning gives a reasonable way to utilize the intrinsic characteristics of unlabeled data to obtain supervisory signals, which has attracted increasing attention.

Different from image data that can be handled by defining proxy tasks (e.g., predicting relative positions of image patches \cite{doersch2015unsupervised}, solving  jigsaw puzzles \cite{noroozi2016unsupervised}, inpainting images \cite{pathak2016context}, and predicting the image color channel \cite{larsson2017colorization}) for self-supervised learning, video data additionally contain temporal information that can be leveraged to learn the supervisory signals. Recently, a variety of approaches have been proposed such as order verification \cite{misra2016shuffle,fernando2017self}, order prediction \cite{lee2017unsupervised, xu2019self,wang2020self}, speediness prediction \cite{benaim2020speednet,yao2020video}. However, these methods consider the temporal dependency only from a single scale (i.e., short-term or long-term) and ignore the multi-scale temporal dependencies, i.e., they extract either snippet-level or frame-level features via 2D/3D CNNs and neglect to integrate these features to model elaborate multi-scale temporal dependencies.

\begin{figure}
\begin{center}
\includegraphics[scale=0.4]{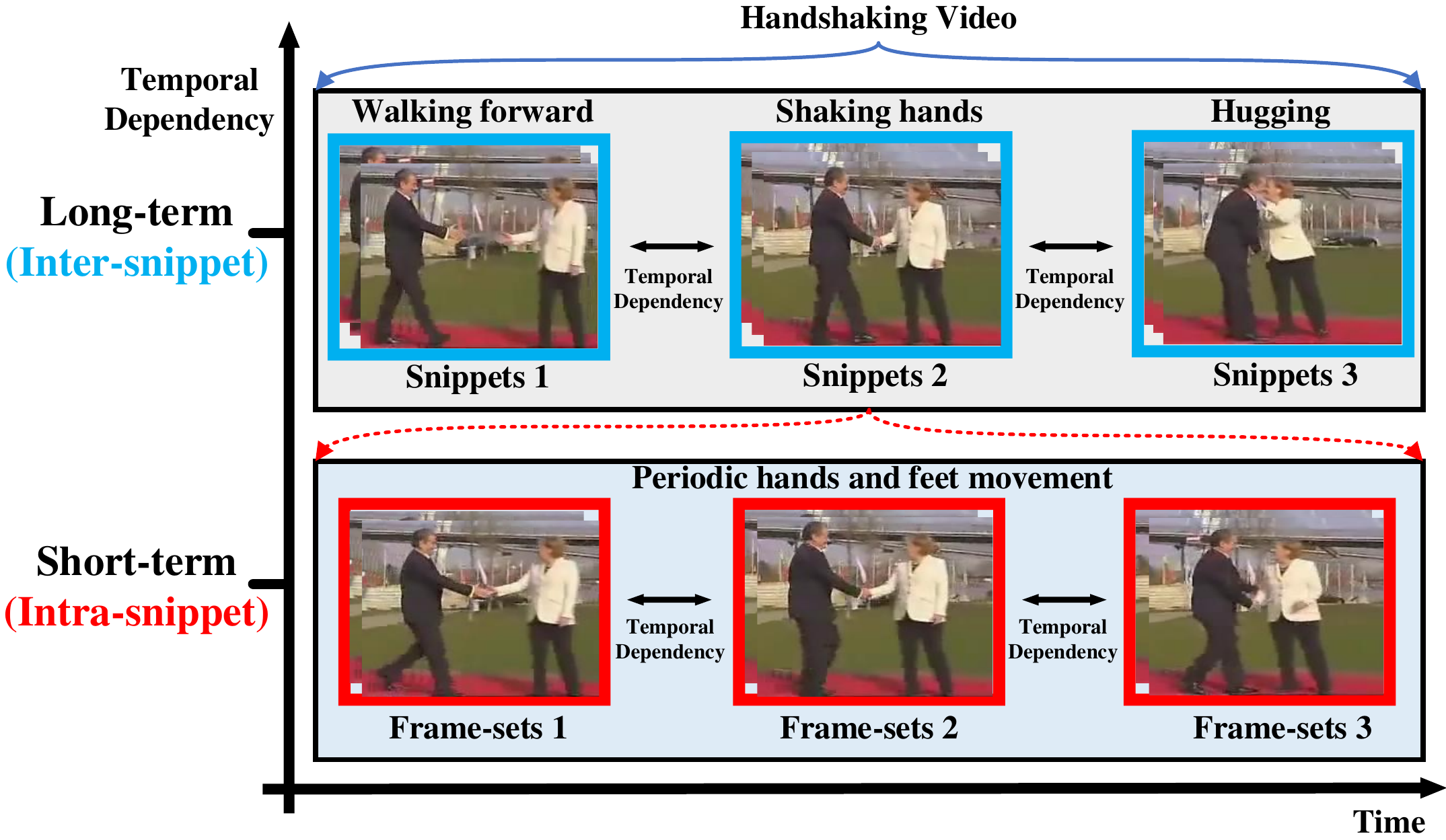}
\end{center}
   \caption{Illustration of the multi-scale temporal dependencies. The handshaking contains the long-term  (inter-snippet) temporal dependencies of walking forward, shaking hands, and hugging, while it also includes the short-term (intra-snippet) temporal dependencies of periodic hands and feet movement. }
\label{Fig1}
\end{figure}

In this work, we argue that modeling multi-scale temporal dependencies is beneficial for various video classification tasks. Firstly, the recent neuroscience studies \cite{livingstone1988segregation,van1994neural, hans2020visual} prove that the human visual system can perceive detailed motion information by capturing both long-term and short-term temporal dependencies. This has been inspired by several famous supervised learning methods (e.g., Nonlocal \cite{wang2018non}, PSANet \cite{zhao2018psanet}, GloRe \cite{chen2019graph}, and ACNet \cite{wang2019adaptively}). Secondly, an action usually consists of several temporal dependencies at both short-term and long-term timescales. Typical actions such as handshaking and drinking, as well as cycles of repetitive actions such as walking and swimming, often last several seconds and span tens or hundreds of frames. In Figure \ref{Fig1}, the video of handshaking contains atomic actions such as walking forward, shaking hands, and hugging, which forms the long-term temporal dependencies (video snippets). Within a snippet, it also includes the short-term temporal dependencies (frame-sets within a snippet) of periodic hands and feet movement. The internal dependency among different snippets and frame-sets contributes to describing the detailed semantic concept for video action analysis. Randomly shuffling the frames or snippets cannot preserve the semantic content of the video. Actually, the short-term temporal dependencies within a video snippet is important especially for videos that contain strict temporal coherence, such as videos in SomethingSomething datasets \cite{goyal2017something}. Therefore, both short-term (e.g., intra-snippet) and long-term (e.g., inter-snippet) temporal dependencies are essential and should be jointly modeled to learn discriminative temporal representations for unlabeled videos. However, long-term and short-term temporal dependencies have complicated interdependence which are correlated and complementary to each other. Inspired by the convincing performance and high interpretability of graph convolutional networks (GCN) \cite{kipf2016semi, velivckovic2018graph, zhang2020deep,yang2020distilling,yang2020factorizable}, several works \cite{wang2018videos,zhuo2019explainable,liu2019social,zhang2019multi,ji2020action,zhang2020temporal} were proposed to increase the temporal diversity by using GCN in a supervised learning fashion with labeled videos. Unfortunately, due to the lack of principles to explore the multi-scale temporal knowledge of unlabeled videos, it is quite challenging to utilize GCN for self-supervised multi-scale temporal dependencies modeling.

\begin{figure}
\begin{center}
\includegraphics[scale=0.44]{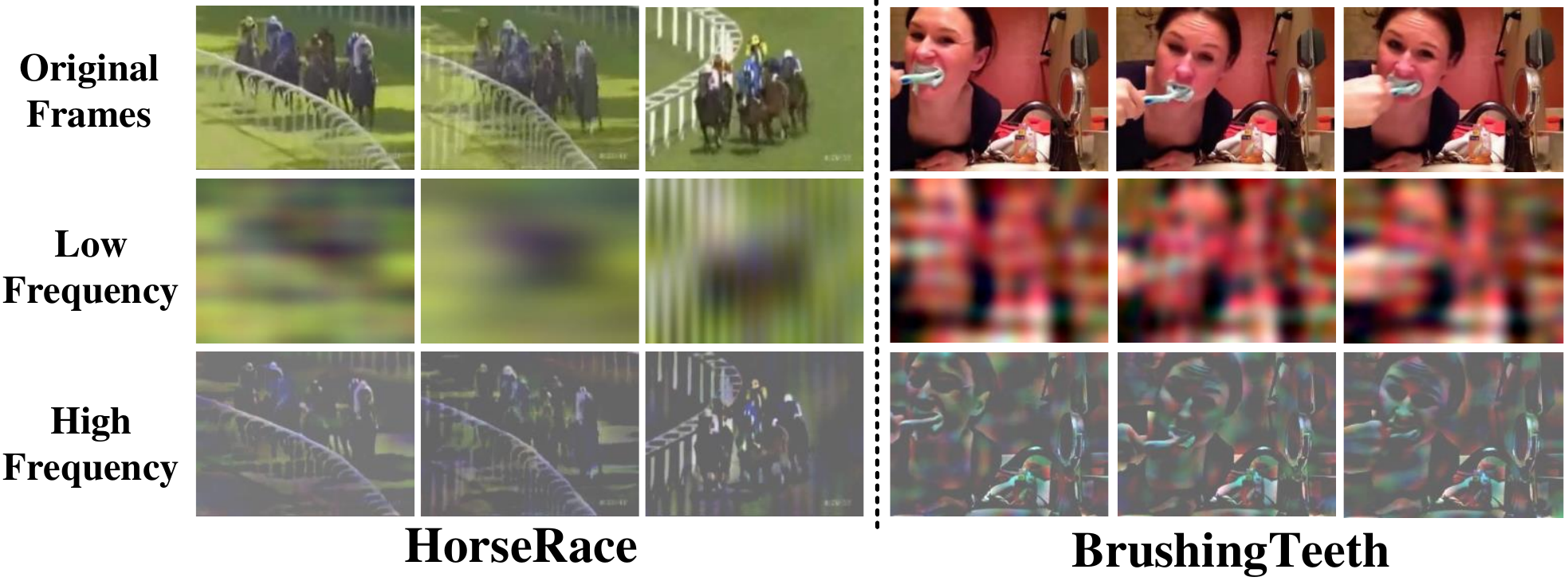}
\end{center}
   \caption{Illustration of the video frames with different frequencies. The first row is the original frames. The second row is the low frequency (focus on scene representation). And the last row is the high frequency (focus on distinct motion edges).}
\label{Fig2}
\end{figure}

For video frames, their inherent information is conveyed at different frequencies \cite{campbell1968application,de1980spatial,chen2019drop}. As shown in Figure \ref{Fig2}, a video frame can be decomposed into a low spatial frequency component that describes the smoothly changing structure (scene representation) and a high spatial frequency that describes the rapidly changing details (motion representation).  The low-frequency representation can retain the most of scene information. While in the high frequency, the scene information would be counteracted, and the distinct motion edges would be highlighted. In order to capture both temporal dynamics and scene appearance through different frequencies, we calculate the feature frequency spectrum along the temporal domain based on the discrete cosine transform, and then distill the discriminative spatial-temporal knowledge from videos.

\textbf{Existing Challenges}. Based on the above observations, the following three key challenges should be addressed: (a)  how to design a specific feature extraction module that highlights motion information and concurrently preserves both spatial and temporal knowledge from videos? (b) how to model multi-scale temporal dependencies in an explicit way to increase the temporal diversity of large amounts of unlabeled videos? (c) how to build an unified self-supervised learning framework that can learn video understanding model with strong video representative power and generate well to downstream tasks?

\textbf{Insights}. To address aforementioned challenges, this work presents a novel video self-supervised learning approach, named Temporal Contrastive Graph Learning (TCGL). The proposed TCGL aims at learning the multi-scale temporal dependency knowledge within videos by guiding the video snippet order prediction in an adaptive manner. Specifically, a given video is sampled into several fixed-length snippets and then randomly shuffled. For each snippet, all the frames from this snippet are sampled into several fixed-length frame-sets. To address challenge (a), we conduct frequency spectrum analysis using Discrete Cosine Transform (DCT), and propose a novel Spatial-Temporal Knowledge Discovering (STKD) module to highlight motion information and fully discover spatial-temporal information within snippets and frame-sets. Then, we utilize 3D CNN as the backbone network to extract features for these motion-enhanced snippets and frame-sets. To address challenge (b), we propose graph neural network (GNN) structures with prior knowledge about the snippet orders and frame-set orders to explicitly model both inter-snippet and intra-snippet temporal dependencies within videos. The video snippets of a video and their temporal dependencies are used to construct the inter-snippet temporal graph. Similarly, the frame-sets within a video snippet and their temporal characteristics are leveraged to construct the intra-snippet temporal graph. To generate different correlated graph views, we randomly remove edges and mask node features of the intra-snippet graphs or inter-snippet graphs. Then, specific contrastive learning modules are designed to enhance its discriminative capability for temporal representation learning. To address challenge (c), we propose a novel pretext task learning module, named Adaptive Snippet Order Prediction (ASOP) module. The ASOP generates supervisory signals for unlabeled videos by adaptively leveraging relational knowledge among video snippets to predict the actual snippet orders. The main contributions of the paper can be summarized as follows:

\begin{itemize}
\item To highlight motion information and discover discriminative spatial-temporal representations from videos, we conduct theoretical analysis in the frequency domain based on discrete cosine transform and propose a novel Spatial-Temporal Knowledge Discovering (STKD) module.
\item Integrated with intra-snippet and inter-snippet temporal dependencies, we propose intra-snippet and inter-snippet Temporal Contrastive Graphs (TCG) to increase the temporal diversity among video frames and snippets in a graph contrastive self-supervised learning manner.
\item To generate supervisory signals for unlabeled videos, we propose a novel pretext task learning module, named Adaptive Snippet Order Prediction (ASOP) module, which leverages relational knowledge among video snippets to predict the actual snippet orders by learning the global context representation and recalibrating the channel-wise features adaptively for each video snippet.
\item Extensive experiments on three networks and two downstream tasks show that the proposed method achieves state-of-the-art performance and demonstrate the great potential of the learned video representations.
\end{itemize}

The rest of the paper is organized as follows. We first review related works in Section 2, the details of the proposed method are then explained in Section 3. In Section 4, the implementation and results of the experiments are provided and analyzed. Finally, we conclude our works in Section 5.

\section{Related Work}
In this section, we will introduce the recent works on supervised video representation learning, self-supervised video representation learning, and frequency domain learning.

\subsection{Supervised Video Representation Learning}
For video representation learning, a large number of supervised learning methods have been received increasing attention. The methods include traditional methods \cite{laptev2005space,klaser2008spatio,wang2013dense,wang2013action,nguyen2014stap,liu2016combining,peng2016bag,liu2018transferable,liu2018hierarchically} and deep learning methods \cite{simonyan2014two,tran2015learning,wang2015action,tran2018closer,wang2018temporal,zhou2018temporal,liu2018global,lin2019tsm,liu2019deep,huang2019bidirectional,9451581,ni2021cross,huang2021novel}. To model and discover temporal knowledge in videos, two-stream CNNs \cite{simonyan2014two} judged the video image and dense optical flow separately, then directly fused the class scores of these two networks to obtain the classification result. C3D \cite{tran2015learning} processed videos with a three-dimensional convolution kernel. Temporal Segment Networks (TSN) \cite{wang2018temporal} sampled each video into several segments to model the long-range temporal structure of videos. Temporal Relation Network (TRN) \cite{zhou2018temporal} introduced an interpretable network to learn and reason about temporal dependencies between video frames at multiple temporal scales. Temporal Shift Module (TSM) \cite{lin2019tsm} shifted part of the channels along the temporal dimension to facilitate information exchanged among neighboring frames. Although these supervised methods achieve promising performance in modeling temporal dependencies, they require large amounts of labeled videos for training an elaborate model, which is time-consuming and labor-intensive.

\subsection{Self-supervised Video Representation Learning}
Although there exists a large amount of videos, it may take a great effort to annotate  such massive data. Self-supervised learning gives a feasible way to generate supervisory signals by modeling various pretext tasks with abundant unlabeled data. The learned model from pretext tasks can be directly applied to downstream tasks for feature extraction or fine-tuning. Specific contrastive learning methods have been proposed, such as the NCE \cite{gutmann2010noise}, MoCo \cite{he2020momentum}, BYOL \cite{grill2020bootstrap}, SimCLR \cite{DBLP:conf/icml/ChenK0H20}. To better model topologies, contrastive learning methods on graphs \cite{qiu2020gcc,zhu2020deep,hafidi2020graphcl,you2020graph} have also been attracted increasing attention. Although the structures of MoCo, SimCLR, BYOL are simple and effective, these methods focus on image recognition without considering the temporal information. Moreover, these methods require large amounts of negative samples or huge batchsize. However, for self-supervised video recognition, videos require much more computational resources than images. Thus, the existing self-supervised image recognition models cannot be directly applied to the video self-supervised learning models due to the limited computational resources. Therefore, we need to elaborately design computationally efficient modules that require less computational resource. Moreover, the videos contain extra temporal information that requires elaborately designed specific modules to explicitly highlight motion information, model temporal dependencies, and generate supervisory signals for unlabeled videos.

For self-supervised video representation learning, how to effectively explore temporal information is important. Many existing works focus on the discovering of temporal information. Shuffle\&Learn \cite{misra2016shuffle} randomly shuffled video frames and trained a network to distinguish whether these video frames are in the right order or not. Odd-one-out Network \cite{fernando2017self} proposed to identify unrelated or odd video clips. Order prediction network (OPN) \cite{lee2017unsupervised} trained networks to predict the correct order of shuffled frames. VCOP \cite{xu2019self} used 3D convolutional networks to predict the orders of shuffled video clips. SpeedNet \cite{benaim2020speednet} designed a network to detect whether a video is playing at a normal rate or sped up rate. Video-pace \cite{wang2020self} utilized a network to identify the right paces of different video clips. In addition to focusing on the temporal dependency, Mas \cite{wang2019self} proposed a self-supervised learning method by regressing both motion and appearance statistics along spatial and temporal dimensions. ST-puzzle \cite{kim2019self} used space-time cubic puzzles to design pretext task. IIC \cite{tao2020self} introduced intra-negative samples by breaking temporal relations in video clips, and used these samples to build an inter-intra contrastive framework. XDC \cite{alwassel2020self} proposed a self-supervised method that leverages unsupervised clustering in audio modality as a supervisory signal for video modality. Though the above works utilize temporal dependency or design specific pretext tasks for video self-supervised learning, the comprehensive temporal diversity and dependency are not fully explored. In our work, we build a novel inter-intra snippet graph structure to model multi-scale temporal dependencies, and produce self-supervision signals about video snippet orders contrastively. Moreover, most of the previous works require stronger models, bigger clip sizes or larger pretrain datasets. While our method can achieve competitive performance with lightweight backbones and smaller clip size.

\begin{figure*}[t]
\begin{center}
\includegraphics[scale=0.37]{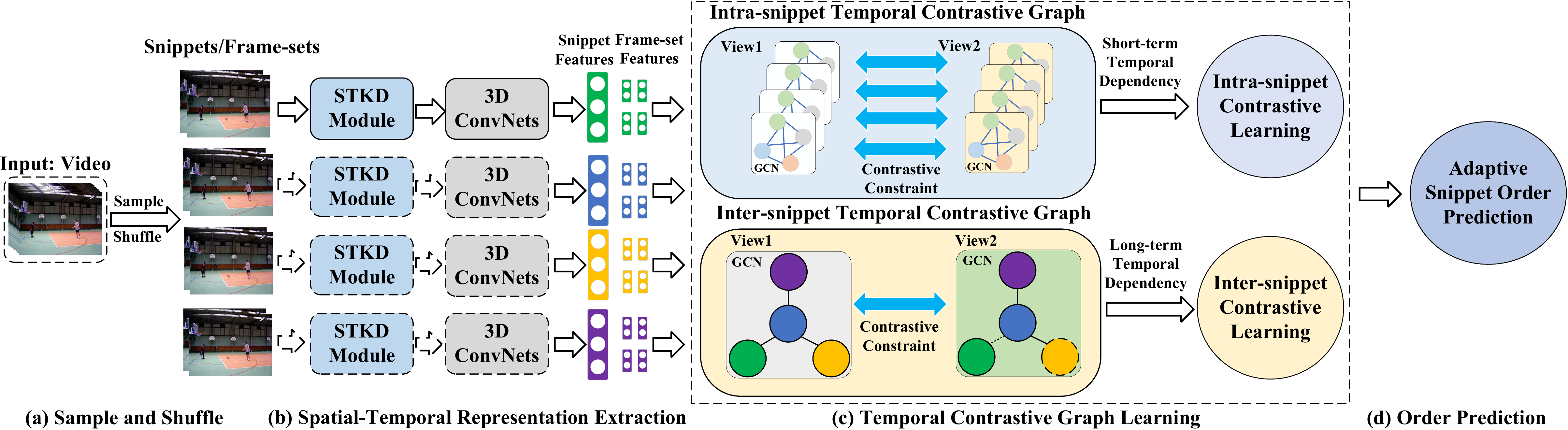}
\end{center}
   \caption{Overview of the TCGL framework. (a) Sample and Shuffle: sample non-overlapping snippets for each video and randomly shuffle their orders. And for each snippets, all the frames from this snippet are sampled into several fixed-length frame-sets. (b) Spatial-Temporal Representation Extraction: discover the discriminative spatial-temporal representation of video snippets and frame-sets by the STKD module. Then, use the 3D CNNs to extract the features for all the motion-enchanced snippets and frame-sets. (c) Temporal Contrastive Graph Learning: intra-snippet and inter-snippet temporal contrastive graphs are constructed with the prior knowledge about the frame-set orders and snippet orders, see Figure \ref{Fig5} for more details. (d) Order Prediction: the learned snippet features from the temporal contrastive graph are adaptively forwarded through an adaptive snippet order prediction module to output the probability distribution over the possible orders.}
\label{Fig3}
\end{figure*}

\subsection{Frequency Domain Learning}
Frequency analysis has always been a powerful tool in the signal processing field. Recently, some frequency analysis methods are proposed in deep learning field. Most of frequency-based methods \cite{jia2018optimizing,wang2020fft} aim to reduce the computing cost and parameters with Fourier Transformation (FT), and therefore improving the network efficiency. Some works \cite{gueguen2018faster,ehrlich2019deep} introduced frequency analysis for JPEG encoding in the CNNs. DCT is introduced in \cite{xu2020learning} to reduce the communication bandwidth. Works in \cite{DBLP:conf/iclr/TorfasonMATTG18,wu2018compressed} proposed dedicated autoencoder-based networks for compression and inference tasks. FcaNet \cite{qin2020fcanet} generalized the pre-processing of channel attention mechanism in the frequency domain. Wang et al., \cite{wang2020towards} applied the analysis of the connection between the distribution of frequency components in the input dataset to conduct the explanation of the CNN. Since the convolution operation in the spatial domain has been proven to be equivalent to the multiplication in the frequency domain \cite{bracewell1986fourier}, \cite{su2020collaborative} performed video knowledge distillation in the frequency domain for action recognition. Frequency Filtering Embedding (FFE) \cite{8778695} used graph Fourier transform and frequency filtering as a graph Fourier domain operator for graph feature extraction. Our method is different from the prior works in two aspects. First, we discover discriminative spatial-temporal knowledge in the frequency domain for video self-supervised learning. Second, we conduct strict theoretical analysis to validate that the discriminative spatial-temporal representation essentially highlights motion information in the high frequency domain.

\section{Temporal Contrastive Graph Learning}

In this section, we first give a brief overview of the proposed method, then clarify each part of the method in detail. We present the overall framework in Figure \ref{Fig3}, which mainly consists of four stages. (1) In sample and shuffle, for each video, several snippets are uniformly sampled and shuffled. For each snippet, all of its frames are sampled into several fixed-length frame-sets. (2) In feature extraction, we discover motion-enhanced spatial-temporal representation of video snippets and frame-sets by the Spatial-Temporal Knowledge Discovering (STKD) module. Then, 3D CNNs are utilized to extract spatial-temporal features for these motion-enhanced snippets and frame-sets, and all 3D CNNs share the same weights. (3) In temporal contrastive learning, we build two kinds of temporal contrastive graph structures (intra-snippet graph and inter-snippet graph) with the prior knowledge about the frame-set orders and snippet orders. To generate different correlated graph views for specific graphs, we randomly remove edges and mask node features of the intra-snippet graphs or inter-snippet graphs. Then, we design specific contrastive losses for both the intra-snippet and inter-snippet graphs to enhance the discriminative capability for discriminative temporal representation learning. (4) In the order prediction, the learned snippet features from the temporal contrastive graph are adaptively forwarded through an adaptive snippet order prediction module to output the probability distribution over the possible orders.

Given a video, the snippets from this video are composed of  frames with the size $c\times l\times h\times w$, where $c$ is the number of channels, $l$ is the number of frames, $h$ and $w$ indicate the height and width of frames. The size of the 3D convolutional kernel is $t\times d\times d$, where $t$ is the temporal length and $d$ is the spatial size. We define an ordered snippet tuples as $\mathbf{S}=\langle s_1, s_2, \cdots, s_n\rangle$, the frame-sets from snippet $s_i $ is denoted as $\mathbf{F}_i=\langle f_1, f_2, \cdots, f_m\rangle$. The subscripts here represent the temporal order. Let $\mathcal{G}=(\mathcal{V},\mathcal{E})$ denote a graph, where $\mathcal{V}=\{v_1,v_2,\cdots,v_N\}$ represents the node set and $\mathcal{E}\in \mathcal{V}\times \mathcal{V}$ represents the edge set. We denote the feature matrix and the adjacency matrix as $\mathbf{X}\in \mathbb{R}^{N\times F}$ and $\mathbf{A}\in \{0,1\}^{N\times N}$, where $\mathbf{x}_i\in \mathbb{R}^F$ is the feature of $v_i$, and $\mathbf{A}_{ij}=1$ if $(v_i,v_j)\in \mathcal{E}$.

\subsection{Sample and Shuffle}
In this stage, we randomly sample consecutive frames (snippets) from the video to construct video snippet tuples. If we sample $N$ snippets from a video, there are $N!$ possible snippet orders. Since the snippet order prediction is purely a proxy task of our TCGL framework and our focus is the learning of 3D CNNs, we restrict the number of snippets of a video between $3$ to $4$ to alleviate the complexity of the order prediction task, inspired by the previous works \cite{noroozi2016unsupervised,xu2019self,xiao2020explore}. The snippets are sampled uniformly from the video with the interval of $p$ frames. After sampling, the snippets are shuffled to form the snippet tuples $\mathbf{S}=\langle s_1, s_2, \cdots, s_n\rangle$. For each snippet $s_i$, all the frames within are uniformly divided into $m$ frame-sets with equal length, then we get the frame-set $\mathbf{F}_i=\langle f_1, f_2, \cdots, f_m\rangle$ for the snippet $s_i$. For snippet tuples, they contain dynamic information and strict temporal dependency of a video, which is essentially the global temporal structure of the videos. For the frame-sets within a snippet, the frame-level temporal dependency among frames provides us the local temporal structure of the videos. By taking both global and local temporal structures into consideration, we can fully explore discriminative temporal dependencies of videos.

\subsection{Discriminative Spatial-temporal Representation Extraction}

\begin{figure}
\begin{center}
\includegraphics[scale=0.59]{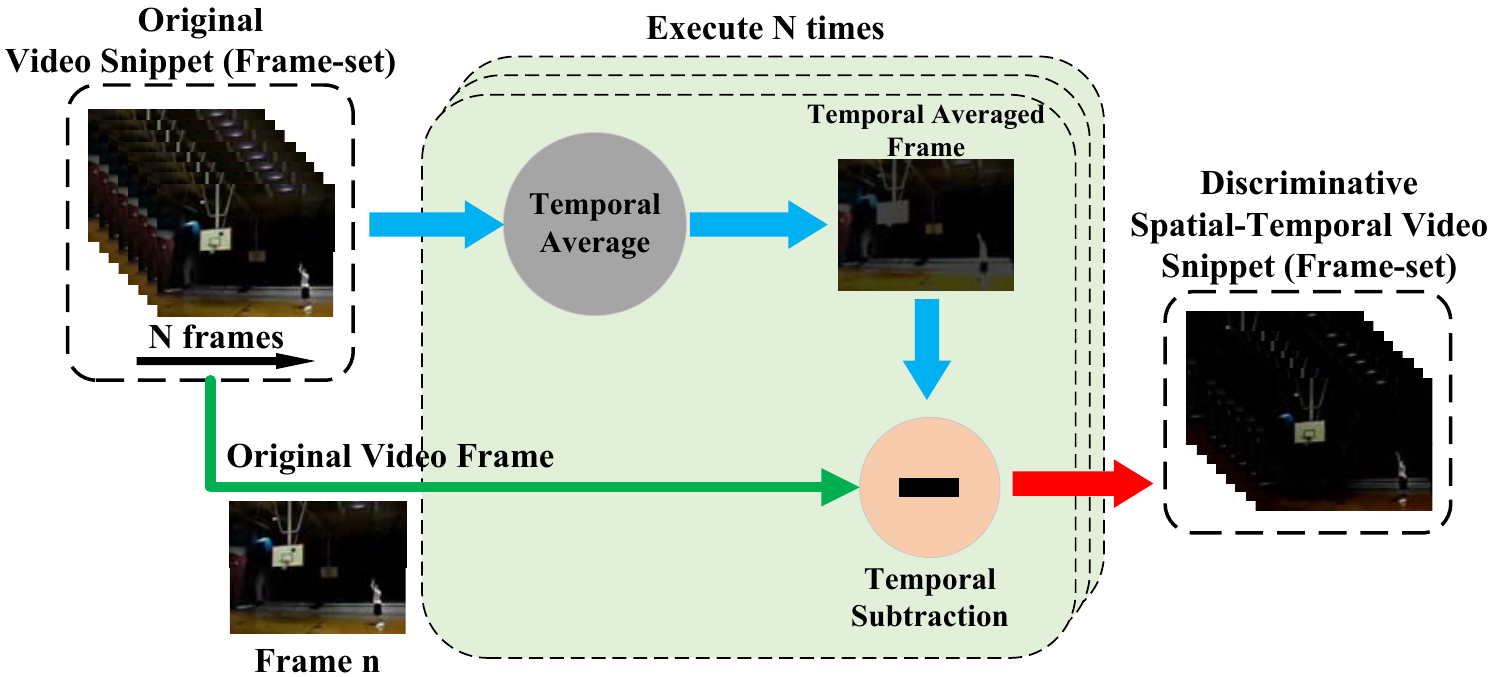}
\end{center}
          \vspace{-10pt}
   \caption{Illustration of the Spatial-Temporal Knowledge Discovering (STKD) module.}
          \vspace{-10pt}
\label{Fig4}
\end{figure}

As shown in Figure \ref{Fig2}, the frequency spectrum computed based on the input video frames can present the motion and scene information with different frequencies. Specifically, high frequency attends to the motion information between neighboring frames, while the low frequency pays attention to the scene representation. Therefore, discriminative spatial-temporal information can be discovered by frequency domain analysis. The overview of the Spatial-Temporal  Knowledge Discovering (STKD) module is shown in Figure \ref{Fig4}. We denote an input video snippet (frame-set) as $\mathbf{V}\in \mathbb{R}^{C\times L\times H\times W}$, where $C$ is the channel number, $L$ denotes the number of the frames, $H$ and $W$ are the width and height of the video frame, respectively. Then we compute its frequency spectrums along the temporal domain and obtain the output features $\mathbf{\hat{T}}\in \mathbb{R}^{C\times K\times H\times W}$, where $K$ is the number of frequency bands. Typically, the definition of  Discrete Cosine Transform (DCT) \cite{ahmed1974discrete} can be written as:
\begin{equation}\label{eq1}
\mathbf{\hat{T}}[k]=\sum\limits_{i=0}^{L-1}\mathbf{V}(i)\textrm{cos}(\frac{2\pi ki}{L}),~ s.t. ~~k\in\{0, 1, \cdots, L-1\}
\end{equation}

Suppose $L=2$ in Eq (\ref{eq1}), we have:
\begin{equation}\label{eq2}
\mathbf{\hat{T}}[0]=\mathbf{V}(0)+\mathbf{V}(1)
\end{equation}
\begin{equation}\label{eq3}
\mathbf{\hat{T}}[1]=\mathbf{V}(0)-\mathbf{V}(1)
\end{equation}

From  Eq (\ref{eq2}), we can find that $\mathbf{\hat{T}}[0]$ represents the low-frequency information, which is the sum of video features, while $\mathbf{\hat{T}}[1]$ denotes the high-frequency information, which is the difference between neighboring video features. Therefore, the low-frequency representation can retain the most of scene information. While in the high frequency, the scene information would be counteracted, and the distinct motion edges would be highlighted. To further validate this phenomenon, we extend $L$ to $3$ and get the following representations:
\begin{equation}\label{eq4}
\begin{aligned}
&\mathbf{\hat{T}}[0]=\mathbf{V}(0)+\mathbf{V}(1)+\mathbf{V}(2)\\
&\mathbf{\hat{T}}[1]=\mathbf{V}(0)-\frac{1}{2}\mathbf{V}(1)-\frac{1}{2}\mathbf{V}(2)\\
&\mathbf{\hat{T}}[2]=\mathbf{V}(0)-\frac{1}{2}\mathbf{V}(1)-\frac{1}{2}\mathbf{V}(2)
\end{aligned}
\end{equation}


To capture both temporal dynamics and scene appearance through different frequencies, we sum over all the frequency components except for the first one $\mathbf{\hat{T}}[0]$ to avoid the influence of low-frequency information when conduct temporal relationship modeling. Thus, we get the discriminative spatial-temporal representations $\widetilde{\mathbf{T}}_2, \widetilde{\mathbf{T}}_3$ for $L=2,3$, respectively:
\begin{equation}\label{eq5}
\begin{aligned}
&\frac{1}{2}\widetilde{\mathbf{T}}_2=\mathbf{V}(0)-\frac{1}{2}(\mathbf{V}(0)+\mathbf{V}(1))\\
&\frac{1}{3}\widetilde{\mathbf{T}}_3=\mathbf{V}(0)-\frac{1}{3}(\mathbf{V}(0)+\mathbf{V}(1)+\mathbf{V}(2))\\
\end{aligned}
\end{equation}

Thus, when $L=n$, we can obtain:
\begin{equation}\label{eq6}
\frac{1}{n}\widetilde{\mathbf{T}}_n=\mathbf{V}(0)-\frac{1}{n}(\mathbf{V}(0)+\mathbf{V}(1)+\cdots+\mathbf{V}(n-1))
\end{equation}

Since $\frac{1}{n}$ in the left of the Eq (\ref{eq6}) is constant, it can be ignored. Thus, Eq (\ref{eq6}) is equivalent to:
\begin{equation}\label{eq7}
\widetilde{\mathbf{T}}_n=\mathbf{V}(0)-\frac{1}{n}(\mathbf{V}(0)+\mathbf{V}(1)+\cdots+\mathbf{V}(n-1))
\end{equation}

Based on the mathematical induction strategy \cite{bather1994mathematical}, we can extend Eq (\ref{eq7}) to its general form when $L=n+1$:
\begin{equation}\label{eq8}
\widetilde{\mathbf{T}}_{n+1}=\mathbf{V}(0)-\frac{1}{n+1}\sum\limits_{i=0}^{n}\mathbf{V}(i)  
\end{equation}

From Eq (\ref{eq8}), we can conclude that the discriminative spatial-temporal representation $\widetilde{\mathbf{T}}_{n}\in \mathbb{R}^{C\times L\times H\times W}$ for a video snippet is essentially the substraction of the original video snippet $\mathbf{V}\in \mathbb{R}^{C\times L\times H\times W}$ with its temporal average pooling $\mathbf{\overline{V}}=\frac{1}{n+1}\sum_{i=0}^{n}\mathbf{V}(i)\in \mathbb{R}^{C\times L\times H\times W}$ along the temporal axis. The STKD module can extract discriminative temporal representation for video snippets without complex calculation and is flexible enough to be inserted into the existing models in plug-and-play manner. Based on these discriminative spatial-temporal representations for video snippets and frame-sets, we choose C3D \cite{tran2015learning}, R3D-18 \cite{tran2018closer} and R(2+1)D-18 \cite{tran2018closer} as feature encoders to further extract spatio-temporal features.



\begin{figure*}
\begin{center}
\includegraphics[scale=0.41]{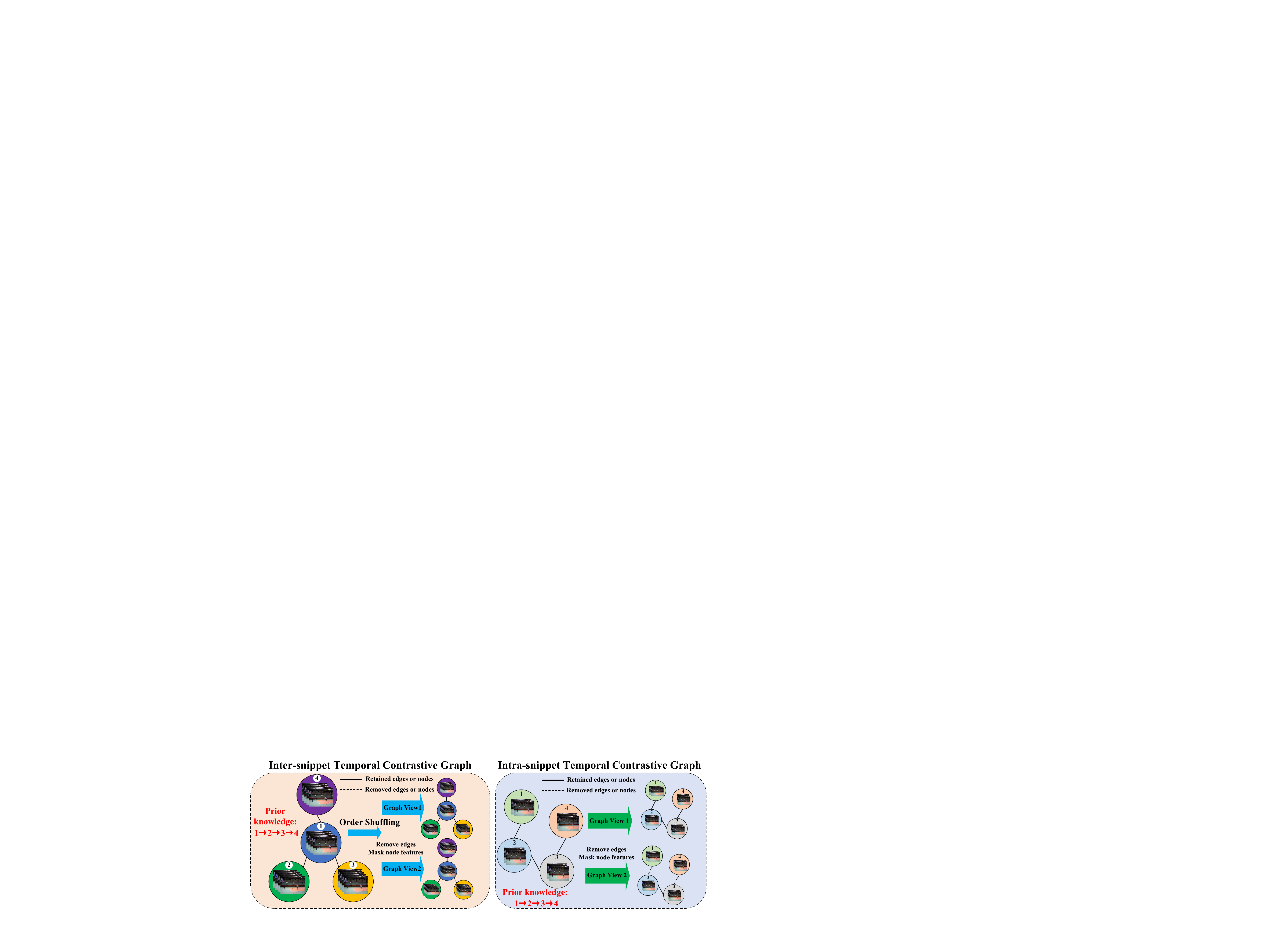}
\end{center}
   \caption{Illustration of inter-snippet and intra-snippet temporal graphs. Inter-snippet and intra-snippet temporal contrastive graphs are constructed with the prior knowledge about the snippet orders  and the frame-set orders of a video.}
\label{Fig5}
\end{figure*}

\subsection{Temporal Contrastive Graph  Learning Module}
Due to the powerful representation ability of graph convolutional network (GCN) \cite{kipf2016semi,zhu2020deep,wu2020comprehensive,yang2020distilling,yang2020factorizable} in complicated relation modeling, the temporal correlation of intra-/inter-snippet features can be explicitly represented by structured graphs, and their dependencies can be captured by adaptive message propagation through the graphs. In addition, the prior temporal relationship can also determine the edges of graphs more explicitly. Therefore, we use it to explore node interaction within each snippet and frame-set for modeling multi-scale temporal dependencies of videos. After obtaining the feature vectors for snippets and frame-sets, we construct two kinds of temporal contrastive graph structures: inter-snippet and intra-snippet temporal contrastive graphs, to increase the temporal diversity of videos, as shown in Figure \ref{Fig3} (c). Assume that a video is shuffled into $4$ snippets, and each snippet is further sampling into $4$ frame-sets, we can obtain corresponding snippets $\mathbf{S}=\langle s_2, s_4, s_1, s_3\rangle$ and the frame-sets $\mathbf{F}_i=\langle f_1, f_2, f_3, f_4\rangle$ for each video snippet $\mathbf{S}_i  (i=1,\cdots,4)$. Based on snippets features $\mathbf{S}$, frame-sets features $\mathbf{F}_i (i=1,\cdots,4)$, and the prior temporal relationship knowledge, we build specific temporal contrastive graphs, as shown in Figure \ref{Fig3} (c).

To build intra-snippet and inter-snippet temporal contrastive graphs, we take advantage of prior knowledge about the temporal relation and the corresponding feature vectors. To fix notation, we denote intra-snippet and inter-snippet graphs as $\mathbf{G}_{\textrm{intra}}^k=\mathcal{G}(\mathbf{X}_{\textrm{intra}}^k,\mathbf{A}_{\textrm{intra}}^k)$ and $\mathbf{G}_{\textrm{inter}}=\mathcal{G}(\mathbf{X}_{\textrm{inter}},\mathbf{A}_{\textrm{inter}})$, respectively, where $k=1,\cdots,m$, $m$ is the number of frame-sets in a video snippet. As shown in Figure \ref{Fig3}, the prior knowledge that the correct order of frames in frame-sets, and the correct order of snippets in snippet tuples are already known because our proxy task is video snippet order prediction. Therefore, we can utilize the prior temporal relationship to determine the edges of graphs. For example, in Figure \ref{Fig5}, if we know that snippets (frame-sets) are ranking temporally $1\rightarrow2\rightarrow3\rightarrow4$, we can connect the temporally related nodes and disconnect the temporally unrelated nodes. To jointly discover the dependencies between inter-snippet and intra-snippet graphs and make graph learning consistent, the $\mathbf{G}_{\textrm{intra}}$ and $\mathbf{G}_{\textrm{inter}}$ share the same structure but different weight parameters. Here, we take inter-snippet temporal graph $\mathbf{G}_{\textrm{inter}}$ as an example to clarify our temporal contrastive graph learning method. Both $\mathbf{G}_{\textrm{inter}}$ and $\mathbf{G}_{\textrm{intra}}^k$ are undirected graphs.

For $\mathbf{G}_{\textrm{inter}}$, we randomly remove edges and masking node features to generate two graph views $\mathbf{\widetilde{G}}_{\textrm{inter}}^1$ and $\mathbf{\widetilde{G}}_{\textrm{inter}}^2$, and the node embeddings of two generated views are denoted as $\mathbf{U}=\mathcal{G}(\mathbf{\widetilde{X}}_{\textrm{inter}}^1,\mathbf{\widetilde{A}}_{\textrm{inter}}^1)$ and $\mathbf{V}=\mathcal{G}(\mathbf{\widetilde{X}}_{\textrm{inter}}^2,\mathbf{\widetilde{A}}_{\textrm{inter}}^2)$. Since different graph views provide different contexts for each node, we corrupt the original graph at both structure and attribute levels to achieve contrastive learning between node embeddings from different views. Therefore, we propose two strategies for generating graph views: removing edges and masking nodes.

The edges in the original graph are randomly removed using a random masking matrix $\mathbf{\widetilde{R}}\in\{0,1\}^{N\times N}$, where the entry of $\mathbf{\widetilde{R}}$ is drawn from a Bernoulli distribution $\mathbf{\widetilde{R}}_{ij}\sim \mathcal{B}(1-p_r)$ if $\mathbf{A}_{ij}=1$ for the original graph, and $\mathbf{\widetilde{R}}_{ij}=0$ otherwise. Here $p_r$ denotes the probability of each edge being removed. Then, the resulting adjacency matrix can be computed as follows, where $\circ$ is Hadamard product.
\begin{equation}\label{eq9}
\mathbf{\widetilde{A}}=\mathbf{A}\circ\mathbf{\widetilde{R}}
\end{equation}
In addition, a part of node features is masked with zeros using a random vector $\mathbf{\widetilde{m}}\in\{0,1\}^F$, where each dimension of it is drawn from a Bernoulli distribution $\mathbf{\widetilde{m}}_i\sim \mathcal{B}(1-p_m), \forall i$. Then, the generated masked features $\mathbf{\widetilde{X}}$ is calculated as follows:
\begin{equation}\label{eq10}
\mathbf{\widetilde{X}}=[\mathbf{x}_1\circ \mathbf{\widetilde{m}};\mathbf{x}_2\circ \mathbf{\widetilde{m}};\cdots; \mathbf{x}_N\circ \mathbf{\widetilde{m}}]^\top
\end{equation}
where $[\cdot;\cdot]$ is the concatenation operator. We jointly leverage these two strategies to generate graph views.

Different from the NCE loss \cite{gutmann2010noise} that only consider inter-view negatives, we take both inter-view and intra-view negative samples into consideration and propose a novel contrastive learning loss that distinguishes embeddings of the same node from these two distinct views from other node embeddings. Given a positive pair, the negative samples come from all other nodes in the two views (inter-view or intra-view). To compute the relationship of embeddings $\mathbf{u}$, $\mathbf{v}$ from two views, we define the relation function $\phi(\mathbf{u},\mathbf{v})=\mathcal{P}(g(\mathbf{u}),g(\mathbf{v}))$, where $\mathcal{P}$ is the L2 normalized dot product similarity, and $g$ is a non-linear projection with two-layer multi-layer perception. The pairwise contrastive objective for positive pair $(\mathbf{u}_i,\mathbf{v}_i)$ is defined as:
\begin{equation}\label{eq11}
\ell(\mathbf{u}_i,\mathbf{v}_i)=\frac{e^{\frac{\phi(\mathbf{u}_i,\mathbf{v}_i)}{\tau}}}{e^{\frac{\phi(\mathbf{u}_i,\mathbf{v}_i)}{\tau}}+\sum\limits_{k=1}^N{\mathbf{\mathbb{I}}_{[k\neq i]}e^{\frac{\phi(\mathbf{u}_i,\mathbf{v}_k)}{\tau}}}+e^{\frac{\phi(\mathbf{u}_i,\mathbf{u}_k)}{\tau}}}
\end{equation}
where $\mathbf{\mathbb{I}}_{[k\neq i]}\in\{0,1\}$ is an indicator function that equals to $1$ if $k\neq i$, and $\tau$ is a temperature parameter, which is empirically set to $0.5$. The first term in the denominator represents the positive pairs, the second term represents the inter-view negative pairs, the third term represents the intra-view negative pairs. Since two views are symmetric, the loss for another view is defined similarly for $\ell(\mathbf{v}_i,\mathbf{u}_i)$. The overall contrastive loss for $\mathbf{G}_{\textrm{inter}}$ is defined as follows:
\begin{equation}\label{eq12}
\mathcal{J}_{\textrm{inter}}=\frac{1}{2N}\sum\limits_{i=1}^N[\ell(\mathbf{u}_i,\mathbf{v}_i)+\ell(\mathbf{v}_i,\mathbf{u}_i)]
\end{equation}

The contrastive loss for intra-snippet graphs $\mathbf{G}_{\textrm{intra}}^k ( k=1,\cdots,m)$ can be computed similarly as $\mathbf{G}_{\textrm{inter}}$.
\begin{equation}\label{eq13}
\mathcal{J}_{\textrm{intra}}^k=\frac{1}{2N}\sum\limits_{i=1}^N[\ell^k(\mathbf{u}_i,\mathbf{v}_i)+\ell^k(\mathbf{v}_i,\mathbf{u}_i)]
\end{equation}

Then, the overall temporal contrastive graph loss is defined as follows, where $\alpha$ and $\beta$ are the weights for intra-snippet graph and inter-snippet graph, respectively.
\begin{equation}\label{eq14}
\mathcal{J}_{g}=\alpha\sum\limits_{k=1}^m\mathcal{J}_{\textrm{intra}}^k+\beta\mathcal{J}_{\textrm{inter}}
\end{equation}

\subsection{Adaptive Order Prediction Module}

\begin{figure}
\begin{center}
\includegraphics[scale=0.52]{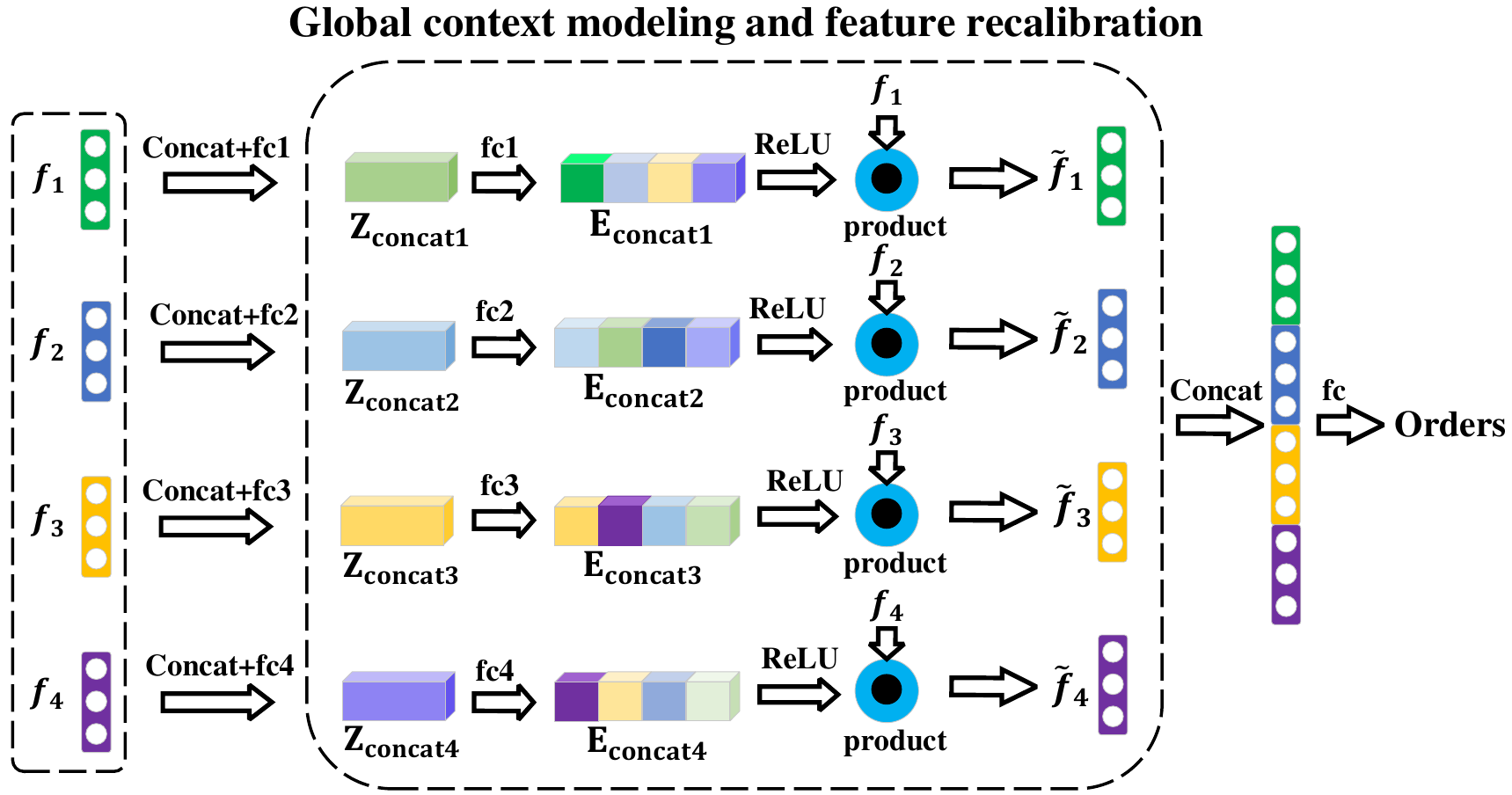}
\end{center}
   \caption{Adaptive snippet order prediction module with four snippets.}
\label{Fig6}
\end{figure}

To generate supervisory signals for unlabeled videos, we propose a novel pretext task learning module, named Adaptive Snippet Order Prediction (ASOP). Since the features from different video snippets are correlated, we build an adaptive order prediction module that receives features from different video snippets and learns a global context embedding, then this embedding is used to recalibrate the input features from different snippets, shown in Figure \ref{Fig6}. We formulate the order prediction task as a classification task using the learned video snippet features from the temporal contrastive graph as the input and the probability distribution of orders as output.

To fix notation, we assume that a video is shuffled into $n$ snippets, the snippet features of nodes learned from inter-snippet temporal contrastive graph are $\{\mathbf{f}_1,\cdots,\mathbf{f}_n\}$, where $\mathbf{f}_k\in \mathbb{R}^{c_k} (k=1,\cdots,n$). To utilize the correlation among these snippets, we concatenate these feature vectors and get joint representations $\mathbf{Z}_u^k$ for each snippet $\mathbf{f}_k$ through a fully-connected layer:
\begin{equation}\label{eq15}
\mathbf{Z}_u^k=\mathbf{W}_{s}^k[\mathbf{f}_1,\cdots,\mathbf{f}_n]+\mathbf{b}_{s}^k, ~~k=1,\cdots,n
\end{equation}
where $[\cdot,\cdot]$ denotes the concatenation operation, $\mathbf{Z}_u^k\in \mathbb{R}^{c_{u}}$ denotes the joint representation, $\mathbf{W}_{s}^k$ and $\mathbf{b}_{s}^k$ are weights and bias of the fully-connected layer. We choose $c_{u}=\frac{\sum_{k=1}^nc_k}{2n}$ to restrict the model capacity and increase its generalization ability. To make use of the global context information aggregated in the joint representations $\mathbf{Z}_{u}^k$, we predict excitation signal for it via a fully-connected layer:
\begin{equation}\label{eq16}
\mathbf{E}^k=\mathbf{W}_{e}^k\mathbf{Z}_u^k+\mathbf{b}_{e}^k, ~~k=1,\cdots,n
\end{equation}
where $\mathbf{W}_{e}^k$ and $\mathbf{b}_{e}^k$ are weights and biases of the fully-connected layer. After obtaining the excitation signal $\mathbf{E}^k\in \mathbb{R}^{c}$, we use it to recalibrate the input feature $\mathbf{f}_k$ adaptively by a simple gating mechanism:
\begin{equation}\label{eq17}
\mathbf{\widetilde{f}}_k=\delta(\mathbf{E}^k)\odot \mathbf{f}_k
\end{equation}
where $\odot$ is channel-wise product operation for each element in the channel dimension, and $\delta(\cdot)$ is the ReLU function. In this way, we can allow the features of one snippet to recalibrate the features of another snippet while concurrently preserving the correlation among different snippets.

Finally, these refined feature vectors $\{\mathbf{\widetilde{f}}_1,\cdots,\mathbf{\widetilde{f}}_n\}$ are concatenated and fed into two-layer perception with soft-max to output the snippet order prediction. To define order classes, we map the snippet orders into different classes. The number of order classes equals to the number of all possible snippet orders. Assume that a video is randomly shuffled into 3 snippets, there are totally 6 (3!) classes. Thus, we can assign class $\{0,1,2,3,4,5\}$ to the snippets according to their orders. For example, when the corresponding snippets is $\{s0, s1, s2\}$, its order class is 0. When the corresponding snippets is $\{s2, s1, s0\}$, its order class is 5.

Then, the cross-entropy loss is used to measure the correctness of the prediction:
\begin{equation}\label{eq18}
\mathcal{J}_{o}=-\sum_{i=1}^{C}\mathbf{y}_i\textrm{log}(\mathbf{p}_i)
\end{equation}
where $\mathbf{y}_i$ and $\mathbf{p}_i$ represent the probability that the sample belongs to the order class $i$ in ground-truth and prediction, respectively. $C$ denotes the number of all possible orders.

The overall self-supervised learning loss for TCGL is obtained by combing Eq (\ref{eq14}) and Eq  (\ref{eq18}), where $\lambda_g$ and $\lambda_o$ control the contribution of $\mathcal{J}_{g}$ and $\mathcal{J}_{o}$  respectively as follow:
\begin{equation}\label{eq19}
\mathcal{J}=\lambda_g\mathcal{J}_{g}+\lambda_o\mathcal{J}_{o}.
\end{equation}

\section{Experiments}
In this section, we first elaborate experimental settings, and then conduct ablation studies to analyze the contribution of key components. Then, the learned 3D CNNs are evaluated on video action recognition and video retrieval tasks with state-of-the-art methods. Finally, we show some visualization results.

\subsection{Experimental Setting}
\textbf{Datasets.} We evaluate our method on three action recognition datasets, UCF101 \cite{soomro2012ucf101}, HMDB51 \cite{kuehne2011hmdb}, Kinetics-400 \cite{kay2017kinetics}, Something-Something-V2 \cite{goyal2017something}, and ActivityNet  \cite{caba2015activitynet}. UCF101 is collected from websites containing $101$ action classes with $9.5$k videos for training and $3.5$k videos for testing. HMDB51 is collected from various sources with $51$ action classes and $3.4$k videos for training and $1.4$k videos for testing. Kinetics-400 (K400) is a large-scale action recognition dataset, which contains 400 action classes and 306k videos. In this work, we use the training split (around 240k videos) as the pre-training dataset. Compared with UCF101, HMDB51 and K400, Something-Something-V2 dataset contains 220,847 videos with 174 classes and focuses more on modeling temporal dependencies. ActivityNet  \cite{caba2015activitynet} is a large-scale human activity understanding benchmark. The latest released version (v1.3) consists of 19,994 videos from 200 activity categories and is utilized here for evaluation. All the videos in the dataset are divided into 10,024, 4,926, and 5,044 for training, validation, and testing sets, respectively. The labels of testing set are not publicly available and thus the performances on ActivityNet dataset are all reported on validation set.

\textbf{Network Architecture.} For video encoder, C3D, R3D-18 and R(2+1)D-18 are used as backbones, where the kernel size of 3D convolutional layers is set to $3\times3\times3$. The R3D-18 network is implemented with no repetitions in conv\{2-5\}\_x, which results in $9$ convolution layers in total. The C3D network is modified by replacing the two fully connected layers with global spatiotemporal pooling layers. The R(2+1)D-18 network has the same architecture as the R3D-18 network with only 3D kernels decomposed. Dropout layers are applied between fully-connected layers with $p=0.1$ for R3D-18, and $p=0.5$ for R3D-50. Our GCN for both inter-snippet and intra-snippet graphs consists of one graph convolutional layer with $512$ output channels.

\textbf{Parameters.} Following the settings in \cite{xu2019self,yao2020video}, we set the snippet length of input video as $16$, the interval length is set as $8$, the number of snippets per tuple is $3$, and the number of frame-sets within each snippet is $4$. During training on UCF101 dataset, we follow the same setting as previous works \cite{xu2019self,xiao2020explore,yao2020video} and randomly split $800$ videos from the training set as the validation set. When training on K400, Something-Something-V2 and ActivityNet datasets, we use the official validation set. Video frames are resized to $128\times171$ ($256\times256$) and then randomly cropped to $112\times112$ ($224\times224$). We set the parameters $\lambda_g=\lambda_o=1$ to balance the contribution between temporal contrastive graph module and adaptive order prediction module. The weights $\alpha$ and $\beta$ are both set to $1$ according to the ablation study result. To optimize the framework, we use mini-bach stochastic gradient descent with the batchsize $16$, the initial learning rate $0.001$, the momentum $0.9$ and the weight decay $0.0005$. The training process lasts for $300$ epochs and the learning rate is decreased to $0.0001$ after $150$ epochs. To make temporal contrastive graphs sensitive to subtle variance between different graph views, the parameters $p_r$ and $p_m$ for generating graph view 1 are empirically set to $0.2$ and $0.1$, and $p_r=p_m=0$ for generating graph view 2. And the values of $p_r$ and $p_m$ are the same for both inter-snippet and intra-snippet graphs. The model with the best validation accuracy is saved to the best model. Our method is implemented by PyTorch \cite{paszke2017automatic} with eight NVIDIA RTX 3090 GPUs.

\textbf{Action Recognition and Video retrieval Protocol.} To make a fair comparison with other methods, we used the linear probe protocol. Specifically, we directly exploit the backbone learnt by TCGL on UCF101 or Kinetics400 datasets as the pretrained model. Then, we finetune the backbones with the pretrained model using the downstream datasets. Only the linear classifier is randomly initialized and updated, other layers are frozen with the pretrained model.

\begin{table}[!t]\renewcommand\tabcolsep{5.0pt}\renewcommand\arraystretch{1}
\begin{center}
\begin{tabular}{lccc}
\hline
Backbone&Snippet Length&Snippets Number&Prediction Accuracy\\\hline
R3D-18&16&2&53.4\\
R3D-18&16&3&\textbf{83.0}\\
R3D-18&16&4&61.1\\\hline
\end{tabular}
\end{center}
\caption{Snippet order prediction accuracy (\%) with different number of snippets within each video.}
\label{Table1}
\end{table}

\begin{table}[!t]\renewcommand\tabcolsep{4.0pt}\renewcommand\arraystretch{1}
\begin{center}
\begin{tabular}{lccc}
\hline
Backbone&Snippet Length&Frame-set Number&Prediction Accuracy\\\hline
R3D-18&16&1&54.1\\
R3D-18&16&2&54.1\\
R3D-18&16&4&\textbf{83.0}\\
R3D-18&16&8&63.1\\\hline
\end{tabular}
\end{center}
\caption{Snippet order prediction accuracy (\%) with various number of frame-sets within each snippet.}
\label{Table2}
\end{table}

\begin{table}[!t]\renewcommand\tabcolsep{3.0pt}\renewcommand\arraystretch{1}
\begin{center}
\begin{tabular}{lcccc}
\hline
Backbone&Snippet Length&Frame-set&GCN Layers&Prediction Accuracy\\\hline
R3D-18&16&4&1&\textbf{83.0}\\
R3D-18&16&4&2&81.1\\
R3D-18&16&4&3&54.6\\\hline
\end{tabular}
\end{center}
\caption{Snippet order prediction accuracy (\%) with different number of GCN layers.}
\label{Table3}
\end{table}

\subsection{Ablation Study}
In this subsection, we conduct ablation studies on the first split of UCF101 with R3D-18 as the backbone, to analyze the contribution of each component of our TCGL and some important hyperparameters. The evaluation metrics are the snippet order prediction accuracy and action recognition accuracy on UCF101 dataset.

\textbf{The number of snippets}. The results of R3D-18 on the snippet order prediction task with different number of snippets are shown in Table \ref{Table1}. The prediction accuracy decreases as the number of snippets increases, due to that the difficulty of the prediction task grows when the snippet number increases. Since more snippets makes the model hard to learn, we use $3$ snippets per video to make a compromise between the task complexity and prediction accuracy.

\textbf{The number of frame-sets}. Since the snippet length is $16$, the number of frame-sets within each snippet can be $1,2,4,8,16$. When the number is $16$, the frame-set only contains static information without temporal information. When the number is $1$ or $2$, it is hard to model the intra-snippet temporal relationship with too few frame-sets. From Table \ref{Table2}, we can observe that more frame-sets within a snippet will make the intra-snippet temporal modeling more difficult, which degrades the order prediction performance. Therefore, we choose $4$ frame-sets per snippet in the experiments for short-term temporal modeling.

\textbf{The number of GCN layers}. From Table \ref{Table3}, we can see that more GCN layers will make model converge more difficultly. Since one GCN layer can achieve comparable performance as two GCN layers, we choose one layer GCN.

\begin{table}[!t]\renewcommand\tabcolsep{5pt}\renewcommand\arraystretch{1}
\begin{center}
\begin{tabular}{lccccc}
\hline
Backbone&Intra ($\alpha$)&Inter ($\beta$)&ASOP&Prediction&Recognition\\\hline
R3D-18&0&0&\Checkmark&55.6&57.3\\
R3D-18&0&1&\Checkmark&55.6&56.0\\
R3D-18&1&0&\Checkmark&76.6&60.9\\
R3D-18&0.1&1&\Checkmark&54.9&56.7\\
R3D-18&1&0.1&\Checkmark&80.2&66.8\\
R3D-18&1&1&\Checkmark&\textbf{83.0}&\textbf{76.8}\\\hline
\end{tabular}
\end{center}
\caption{Snippet order prediction and action recognition accuracy (\%) with different values of $\alpha$ and $\beta$.}
\label{Table4}
\end{table}

\begin{table}[!t]\renewcommand\tabcolsep{4.0pt}\renewcommand\arraystretch{1}
\begin{center}
\begin{tabular}{lccccc}
\hline
Backbone&Intra ($\alpha$)&Inter ($\beta$)&ASOP&Prediction&Recognition\\\hline
R3D-18&1&1&\XSolidBrush&78.4&67.6\\
R3D-18&1&1&\Checkmark&\textbf{83.0}&\textbf{76.8}\\\hline
\end{tabular}
\end{center}
\caption{Snippet order prediction and action recognition accuracy (\%) with/without ASOP module.}
\label{Table5}
\end{table}

\begin{table}[!t]\renewcommand\tabcolsep{4.0pt}\renewcommand\arraystretch{1}
\begin{center}
\begin{tabular}{lccccc}
\hline
Method&Backbone&Pretrain Dataset&STKD&UCF101&HMDB51\\\hline
TCGL&C3D&UCF101&\XSolidBrush&69.5&35.1\\
TCGL&C3D&UCF101&\Checkmark&\textbf{77.4}&\textbf{39.5}\\\hline
TCGL&C3D&K400&\XSolidBrush&75.2&38.9\\
TCGL&C3D&K400&\Checkmark&\textbf{77.9}&\textbf{43.0}\\\hline
TCGL&R3D-18&UCF101&\XSolidBrush&67.6&30.8\\
TCGL&R3D-18&UCF101&\Checkmark&\textbf{76.8}&\textbf{38.8}\\\hline
TCGL&R3D-18&K400&\XSolidBrush&76.6&39.5\\
TCGL&R3D-18&K400&\Checkmark&\textbf{77.6}&\textbf{41.5}\\\hline
TCGL&R(2+1)D-18&UCF101&\XSolidBrush&74.9&36.2\\
TCGL&R(2+1)D-18&UCF101&\Checkmark&\textbf{77.7}&\textbf{40.1}\\\hline
TCGL&R(2+1)D-18&K400&\XSolidBrush&77.6&39.7\\
TCGL&R(2+1)D-18&K400&\Checkmark&\textbf{78.5}&\textbf{41.4}\\\hline
\end{tabular}
\end{center}
\caption{Action recognition accuracy (\%) with/without spatial-temporal knowledge discovering (STKD) module.}
\label{Table6}
\end{table}

\begin{table}[!t]\renewcommand\tabcolsep{4.0pt}\renewcommand\arraystretch{1}
\begin{center}
\begin{tabular}{lcccccccc}
\hline
Backbone&Intra ($\alpha$)&Inter ($\beta$)&ASOP&$\lambda_g$&$\lambda_o$&Prediction Accuracy\\\hline
R3D-18&1&1&\Checkmark&0&1&77.7\\
R3D-18&1&1&\Checkmark&1&0&21.7\\
R3D-18&1&1&\Checkmark&1&0.1&78.9\\
R3D-18&1&1&\Checkmark&0.1&1&83.0\\
R3D-18&1&1&\Checkmark&1&1&\textbf{83.6}\\\hline
\end{tabular}
\end{center}
\caption{Snippet order prediction accuracy (\%) with different values of $\lambda_g$ and $\lambda_o$.}
\label{Table7}
\end{table}

\begin{table}[!t]\renewcommand\tabcolsep{2.0pt}\renewcommand\arraystretch{1}
\begin{center}
\begin{tabular}{lccc}
\hline
Backbone&View Generating methods&Prediction&Recognition\\\hline
R3D-18&Add Random Gaussian Noise&56.1&51.8\\
R3D-18&Randomly Remove Edges&55.2&53.4\\
R3D-18&Randomly Remove Nodes&82.9&74.0\\
R3D-18&Randomly Remove Edges and Nodes&\textbf{83.0}&\textbf{76.8}\\\hline
\end{tabular}
\end{center}
\caption{Snippet order prediction and action recognition accuracy (\%) with different methods of generating views.}
\label{Table8}
\end{table}

\textbf{The intra-snippet and inter-snippet graphs}. To analyze the contribution of intra-snippet and inter-snippet temporal contrastive graphs, we set different values to $\alpha$ and $\beta$ in Eq (\ref{eq14}), shown in Table \ref{Table4}. To be noticed, removing intra-snippet graph will degrade the performance significantly even with the inter-snippet graph, which verifies the importance of intra-snippet graphs for modeling short-term temporal dependency. When setting the weight values of intra-snippet graph and inter-snippet graph to $1$ and $0.1$, the prediction accuracy is $80.2\%$. After exchanging their weight values, the accuracy drops to $54.9\%$. This validates that short-term temporal dependency should be attached more importance. In addition, the performance of TCGL drops significantly when removing either or both of the graphs. When $\alpha=\beta=1$, the prediction accuracy is the best ($83.0\%$). These results validate that both intra-snippet and inter-snippet temporal contrastive graphs are essential for increasing the temporal diversity of features.

\textbf{The adaptive snippet order prediction}. To analyze the contribution of our proposed adaptive snippet order prediction (ASOP) module, we remove this module and merely feed the concatenated features into multi-layer perception with soft-max to output the final snippet order prediction. It can be observed in Table \ref{Table5} that our TCGL performs better than the TCGL without the ASOP module in both order prediction and action recognition tasks. This verifies that the ASOP module can better utilize relational knowledge among video snippets than simple concatenation.

\textbf{The spatial-temporal knowledge discovering}. To analyze the contribution of our proposed spatial-temporal knowledge discovering (STKD) module, we remove this module and merely feed the raw video frames into the backbones. It can be observed in Table \ref{Table6} that our TCGL performs better than the TCGL without the STKD module across all backbones and datasets. This verifies that the STKD module can discover more discriminative spatial-temporal representations for self-supervised video representation learning, and generalizes well to different backbones and datasets.

\textbf{The values of $\lambda_g$ and $\lambda_o$}. Table \ref{Table7} shows that the performance can drop significantly without either TCG loss or ASOP loss. This validates the importance of both TCG and ASOP modules. The performance is the best when $\lambda_g=\lambda_o=1$, which shows that we should attach the same importance to graph contrastive learning and snippet order prediction.

\textbf{The methods of generating views}. Actually, how to create graph views is still an open problem. However, randomly masking out snippets neglects the topology among samples. To produce different graph views at both topology and feature levels, we propose an effective strategy that randomly remove edges/nodes. To justify our superiority on modeling both feature and topology levels, we make comparisons with three methods of generating views: (1) adding random Gaussian noise; (2) randomly removing edges on graphs; (3) randomly removing nodes on graphs, as shown in Table \ref{Table8}. The results in the third row are much better than those in the second row. Actually, the multi-scale temporal dependencies of videos can be explicitly represented by adaptive node message propagation through the graphs. The edges of intra-snippet and inter-snippet temporal contrastive graphs denotes the prior temporal relationship. Therefore, we have the following observations:

(1) Randomly removing the nodes means that the contrastive learning is conduct on sequence of nodes without changing the original temporal graph structure. Therefore, the order prediction task can still work well for these corrupted nodes.

(2) Since the edges of intra-snippet and inter-snippet temporal contrastive graphs denotes the prior temporal relationship, randomly removing them may change the original temporal dependencies of snippet nodes and thus affect the order prediction accuracy.

(3) However, only randomly masking out snippet nodes neglects the topology among samples. Therefore, we need to produce different graph views at both topology and feature levels. In our TCGL, the proposed hybrid graph contrastive learning strategy can fully utilize the complementary characteristics of feature-level and topology-level graph contrastive learning modules, and make them work collaboratively. To be noticed, the fourth row achieves the best performance. This validates that the feature-level and topology-level graph contrastive learning are complementary and can work collaboratively. Results in Table \ref{Table8} justify our superiority on modeling both feature and topology levels.

\subsection{Action Recognition}

\begin{table}[!t]\renewcommand\tabcolsep{0.5pt}\renewcommand\arraystretch{1}
\begin{center}
\begin{tabular}{lccccc}
\hline
Method&Backbone&Pretrain &Clip Size&UCF101&HMDB51\\\hline
Object Patch \cite{wang2015unsupervised}&AlexNet&UCF101&227$\times$227&42.7&15.6 \\
Shuffle \cite{misra2016shuffle}&CaffeNet&UCF101&224$\times$224&50.9&19.8 \\
OPN \cite{lee2017unsupervised}&VGG&UCF101&80$\times$80&56.3&22.1 \\
Deep RL \cite{buchler2018improving}&CaffeNet&UCF101&227$\times$227&58.6&25.0 \\
MoCo \cite{he2020momentum}&I3D&UCF101&16$\times$224$\times$224&70.4&36.3 \\
MemDPC \cite{han2020memory}&R2D3D-34&K400&40$\times$224$\times$224&78.1&41.2 \\
SpeedNet \cite{benaim2020speednet}&S3D&K400&32$\times$224$\times$224&81.1&48.8 \\
XDC \cite{alwassel2020self} &R(2+1)D-50&K400&32$\times$224$\times$224&86.8&52.6\\
SeCo \cite{yao2020seco}&R-50&K400&$224\times224$&88.2&55.5\\
CORP \cite{hu2021contrast}&I3D&K400&32$\times$224$\times$224&90.2&58.7\\
ELo \cite{piergiovanni2020evolving}&R(2+1)D-50&Youtube-8M&N/A&84.2&53.7\\\hline\hline
Random (Baseline)&C3D&-&16$\times$112$\times$112&61.8&24.7\\
Mas \cite{wang2019self}&C3D&UCF101&16$\times$112$\times$112&58.8&32.6 \\
MoCo \cite{he2020momentum}&C3D&UCF101&16$\times$112$\times$112&60.5&27.2 \\
VCOP\cite{xu2019self}&C3D&UCF101&16$\times$112$\times$112&65.6&28.4\\
COP\cite{xiao2020explore}&C3D&UCF101&16$\times$112$\times$112&66.9&31.8\\
PRP \cite{yao2020video}&C3D&UCF101&16$\times$112$\times$112&69.1&34.5\\
STS \cite{wang2021self}&C3D&UCF101&16$\times$112$\times$112&69.3&34.2\\
RTT \cite{RTT}&C3D&UCF101&16$\times$112$\times$112&68.3&38.4\\
\textbf{TCGL (Ours)}&C3D&UCF101&16$\times$112$\times$112&\underline{77.4}&\underline{39.5}\\
\textbf{TCGL (Ours)}&C3D&UCF101&16$\times$224$\times$224&\textbf{79.5}&\textbf{44.4}\\\hline
ST-puzzle \cite{kim2019self}&C3D&K400&16$\times$112$\times$112&60.6&28.3 \\
Mas \cite{wang2019self}&C3D&K400&16$\times$112$\times$112&61.2&33.4\\
CVRL \cite{qian2021spatiotemporal}&C3D&K400&32$\times$112$\times$112&69.9&39.6\\
RTT \cite{RTT}&C3D&K400&16$\times$112$\times$112&69.9&39.6\\
STS \cite{wang2021self}&C3D&K400&16$\times$112$\times$112&71.8&37.8\\
RSPNet \cite{chen2020rspnet}&C3D&K400&16$\times$112$\times$112&76.7&\underline{44.6}\\
\textbf{TCGL (Ours)}&C3D&K400&16$\times$112$\times$112&\underline{77.9}&43.0\\
\textbf{TCGL (Ours)}&C3D&K400&16$\times$224$\times$224&\textbf{81.0}&\textbf{49.2}\\\hline\hline
Random (Baseline)&R3D-18&-&16$\times$112$\times$112&54.5&23.4\\
VCOP \cite{xu2019self}&R3D-18&UCF101&16$\times$112$\times$112&64.9&29.5\\
COP \cite{xiao2020explore}&R3D-18&UCF101&16$\times$112$\times$112&66.0&28.0\\
TCP \cite{lorre2020temporal}&R3D-18&UCF101&10$\times$224$\times$224&64.8&34.7\\
PRP \cite{yao2020video}&R3D-18&UCF101&16$\times$112$\times$112&66.5&29.7\\
STS \cite{wang2021self}&R3D-18&UCF101&16$\times$112$\times$112&67.2&32.7\\
IIC \cite{tao2020self}&R3D-18&UCF101&16$\times$112$\times$112&74.4&38.3\\
RTT \cite{RTT}&R3D-18&UCF101&16$\times$112$\times$112&77.3&\textbf{47.5}\\
\textbf{TCGL (Ours)}&R3D-18&UCF101&16$\times$112$\times$112&76.8&38.8\\
\textbf{TCGL (Ours)}&R3D-18&UCF101&16$\times$224$\times$224&\underline{78.5}&39.4\\
\textbf{TCGL (Ours)}&R3D-50&UCF101&16$\times$112$\times$112&77.9&39.5\\
\textbf{TCGL (Ours)}&R3D-50&UCF101&16$\times$224$\times$224&\textbf{79.7}&\underline{40.4}\\\hline
ST-puzzle \cite{kim2019self}&R3D-18&K400&16$\times$112$\times$112&65.8&33.7 \\
DPC \cite{han2019video}&R3D-18&K400&16$\times$128$\times$128&68.2&34.5\\
DPC \cite{han2019video}&R3D-18&K400&16$\times$224$\times$224&75.7&35.7\\
TCP \cite{lorre2020temporal}&R3D-18&K400&10$\times$224$\times$224&70.5&41.1\\
RSPNet \cite{chen2020rspnet}&R3D-18&K400&16$\times$112$\times$112&74.3&41.8\\
VideoMoCo \cite{pan2021videomoco}&R3D-18&K400&32$\times$112$\times$112&74.1&43.6\\
RTT \cite{RTT}&R3D-18&K400&16$\times$112$\times$112&79.3&\textbf{49.8}\\
\textbf{TCGL (Ours)}&R3D-18&K400&16$\times$112$\times$112&77.6&41.5\\
\textbf{TCGL (Ours)}&R3D-18&K400&16$\times$224$\times$224&\underline{80.3}&42.8\\
\textbf{TCGL (Ours)}&R3D-50&K400&16$\times$112$\times$112&78.8&42.0\\
\textbf{TCGL (Ours)}&R3D-50&K400&16$\times$224$\times$224&\textbf{81.5}&\underline{43.7}\\\hline\hline
Random (Baseline)&R(2+1)D-18&-&16$\times$112$\times$112&55.8&22.0\\
VCP \cite{luo2020video}&R(2+1)D-18&UCF101&16$\times$112$\times$112&66.3&32.2\\
VCOP \cite{xu2019self}&R(2+1)D-18&UCF101&16$\times$112$\times$112&72.4&30.9\\
STS \cite{wang2021self}&R(2+1)D-18&UCF101&16$\times$112$\times$112&73.6&34.1\\
COP \cite{xiao2020explore}&R(2+1)D-18&UCF101&16$\times$112$\times$112&74.5&34.8\\
PRP \cite{yao2020video}&R(2+1)D-18&UCF101&16$\times$112$\times$112&72.1&35.0\\
V-pace \cite{wang2020self}&R(2+1)D-18&UCF101&16$\times$112$\times$112&75.9&35.9\\
PSP \cite{cho2020self}&R(2+1)D-18&UCF101&16$\times$112$\times$112&74.8&36.8\\
\textbf{TCGL (Ours)}&R(2+1)D-18&UCF101&16$\times$112$\times$112&\underline{77.7}&\underline{40.1}\\
\textbf{TCGL (Ours)}&R(2+1)D-18&UCF101&16$\times$224$\times$224&\textbf{79.1}&\textbf{47.5}\\\hline
V-pace \cite{wang2020self}&R(2+1)D-18&K400&16$\times$112$\times$112&77.1&36.6\\
STS \cite{wang2021self}&R(2+1)D-18&K400&16$\times$112$\times$112&77.8&40.7\\
VideoMoCo \cite{pan2021videomoco}&R(2+1)D-18&K400&32$\times$112$\times$112&78.7&\underline{49.2}\\
RSPNet \cite{chen2020rspnet}&R(2+1)D-18&K400&16$\times$112$\times$112&81.1&44.6\\
RTT \cite{RTT}&R(2+1)D-18&K400&16$\times$112$\times$112&\textbf{81.6}&46.4\\
\textbf{TCGL (Ours)}&R(2+1)D-18&K400&16$\times$112$\times$112&78.5&41.4\\
\textbf{TCGL (Ours)}&R(2+1)D-18&K400&16$\times$224$\times$224&\underline{81.2}&\textbf{50.1}\\\hline\hline
\end{tabular}
\end{center}
\vspace{-10pt}
\caption{Comparison with the state-of-the-art self-supervised learning methods on UCF101 and HMDB51. The best and the second-best performance are marked in bold and underline styles, respectively. }
\vspace{-20pt}
\label{Table9}
\end{table}

\subsubsection{Performance on UCF101 and HMDB51}
To verify the effectiveness of our TCGL in action recognition, we initialize the backbones with the model pretrained on the first split of UCF101 or the whole K400 training-set, and fine-tune on UCF101 and HMDB51, the fine-tuning stops after $150$ epochs. The features extracted by the backbones are fed into fully-connected layers to obtain the prediction. For testing, we sample $10$ clips for each video to generate clip predictions, and then average these predictions to obtain the final prediction results. The average classification accuracy over three splits is reported and compared with other self-supervised methods in Table \ref{Table9}, where the backbones, pretrained datasets, and input clip size are illustrated as well. The ``Random" means the model is randomly initialized without pre-training.

From Table \ref{Table9}, we can have the following observations: (1) With the same evaluation metric, our TCGL performs favorably against existing approaches under the C3D, R3D-18 and R(2+1)D-18 backbones on both UCF101 and HMDB51 datasets. (2) Compared with random initialization, our TCGL achieves significant improvement on both UCF101 and HMDB51 datasets, which demonstrates the great potential of our TCGL in self-supervised video representation learning. (3) After pre-trained with the C3D backbone, we outperform the current best-performing methods RTT \cite{RTT} on both UCF101 and HMDB51 datasets. Although the RTT performs slightly better than our TCGL under the R3D-18 and R(2+1)D-18 backbones, our TCGL can achieve more stable performance on UCF101 across three backbones (MAP across three backbones: 78.0\%)  than the RTT (MAP across three backbones: 76.9\%). This validates our good generalization ability for different backbones. In addition, we consistently outperform the other state-of-the-art methods for nearly all evaluation metrics. This validates the scalability and effectiveness of the TCGL. (4) When pre-trained on UCF101 dataset, we achieve better accuracies than some K400 pre-trained methods (Mas \cite{wang2019self}, ST-puzzle \cite{kim2019self} and V-pace \cite{wang2020self}). This verifies that modeling multi-scale temporal dependencies can fully discover temporal knowledge of limited unlabeled videos. (5) Although the recent published works SpeedNet \cite{benaim2020speednet}, XDC \cite{alwassel2020self} and ELo \cite{piergiovanni2020evolving} perform better than our TCGL, they require stronger backbones (e.g. S3D, R(2+1)D-50, R2D3D-34), bigger clip size (e.g. 32$\times$224$\times$224), larger pretrain dataset (e.g. Youtube-8M) and bigger batchsize (e.g. 512). While our TCGL can achieve competitive performance with lightweight backbones (C3D, R3D-18, R(2+1)D-18), smaller clip size (16$\times$112$\times$112) and batchsize (16), which is more computationally efficient and require less computational resource. (6) With lightweight backbone and smaller clip size, we achieve comparable accuracy with the MemDPC \cite{han2020memory}. (7) When adopting bigger clip size (16$\times$224$\times$224), our TCGL can achieve significant performance improvement. For example, with R(2+1)D-18 pretrained on K400, we achieve 81.2\% and 50.1\% on UCF101 and HMDB51, respectively. (8) The TCGL with R3D-50 backbone can obtain nearly 1\% and 0.5\% performance gains than that of the R3D-18 backbone on UCF101 and HMDB51 datasets, respectively. This validates that the R3D-50 backbone is more powerful than the R3D-18 for spatial-temporal representation learning. These observations validate the advantages of our TCGL in learning discriminative spatial-temporal representations for self-supervised video action recognition.

\subsubsection{Performance on Something-Something-V2}
To verify the effectiveness of our TCGL in action recognition scenario which requires strict temporal modeling, we initialize the backbones with the model pretrained the whole K400 training-set, and then fine-tune on Something-Something-V2. The fine-tuning stops after $150$ epochs. The features extracted by the backbones are fed into fully-connected layers to obtain the prediction. For the w/o pretraining methods, the models are randomly initialized. For the fully supervised methods, the C3D and R3D-18 are fully trained on K400 dataset, which is a large-scale dataset with manually annotated action labels, and thus the supervised pre-trained models are strong baselines for our unsupervised pre-trained TCGL. In Table \ref{Table10}, despite not using manual annotations, TCGL increases the accuracy compared with the random initialized model. Surprisingly, with C3D and R3D-18, our TCGL even outperforms the supervised pretrained model, increasing from (47.0\%, 43.7\%) to (48.5\%, 44.6\%), respectively. In addition, we also perform better than the self-supervised method RSPNet \cite{chen2020rspnet} under the same setting across C3D and R3D-18 backbones. These results show that our TCGL can learn discriminative spatial-temporal representations on action recognition dataset that requires strict temporal modeling, which validates the great potential of our TCGL in self-supervised video understanding.

\begin{table}[t]\renewcommand\tabcolsep{1pt}\renewcommand\arraystretch{1}\renewcommand\tabcolsep{25.0pt}
\begin{center}
\begin{tabular}{lcc}
\hline
Method&C3D&R3D-18\\\hline
W/O pre-training&45.8&42.1 \\
Fully supervised&47.0&43.7 \\
RSPNet \cite{chen2020rspnet}&47.8&44.0 \\
\textbf{TCGL (Ours)}&\textbf{48.5}&\textbf{44.6}\\\hline
\end{tabular}
\end{center}
\caption{Top-1 action recognition performance on Something-Something-V2 validation set.}
\label{Table10}
\end{table}

\subsubsection{Performance on Kinetics400 and ActivityNet datasets}
To verify the effectiveness of our TCGL on downstream tasks of action recognition and untrimmed activity recognition, we follow the settings of  \cite{yao2020seco} and pretrain the R3D-50 backbone by TCGL on Kinetics400, and then finetune the pretrained R3D-50 backbone with the downstream datasets Kinetics400 and ActivityNet, respectively. All layers except the last linear layer are frozen with the pre-training backbone. The finetuned model is exploited as the feature extractor to verify the frozen representation via linear classification. For each video in Kinetics400 and ActivityNet, we uniformly sample 30 and 50 frames, respectively, resize each frame with short edge of 256, and crop the resized version to $224\times224$ by using center crop. The linear classifier is finally trained on the training videos of Kinetics400 or ActivityNet and evaluated on each validation set. We adopt the top-1 accuracy as the performance metric, as shown in Table \ref{Table11}.

\begin{table}[t]\renewcommand\tabcolsep{1pt}\renewcommand\arraystretch{1}\renewcommand\tabcolsep{0.6pt}
\begin{center}
\begin{tabular}{lccccc}
\hline
Method&Backbone&Pretrain&Batchsize&K400&ActivityNet\\\hline
MoCo-ImageNet&R-50&ImageNet&512&51.3&66.1\\
ImageNet Pre-training&R-50&ImageNet&512&52.3&67.1\\\hline
SeCo-Inter \cite{yao2020seco}&R-50&ImageNet+K400&512&58.9&66.6 \\
SeCo-Inter+Intra \cite{yao2020seco}&R-50&ImageNet+K400&512&60.7&68.3 \\
SeCo \cite{yao2020seco}&R-50&ImageNet+K400&512&61.9&\textbf{68.5} \\
$\textrm{CORP}_m$ \cite{hu2021contrast}&R3D-50&K400&64&59.1&- \\
$\textrm{CORP}_f$ \cite{hu2021contrast}&R3D-50&K400&512&\textbf{66.3}&- \\
\textbf{TCGL (Ours)}&R3D-50&K400&16&57.6&61.8\\\hline
\end{tabular}
\end{center}
\caption{Top-1 linear evaluation results of the proposed TCGL on the Kinetics-400 and ActivityNet datasets.}
\label{Table11}
\end{table}

\begin{figure*}
\begin{center}
\includegraphics[scale=0.45]{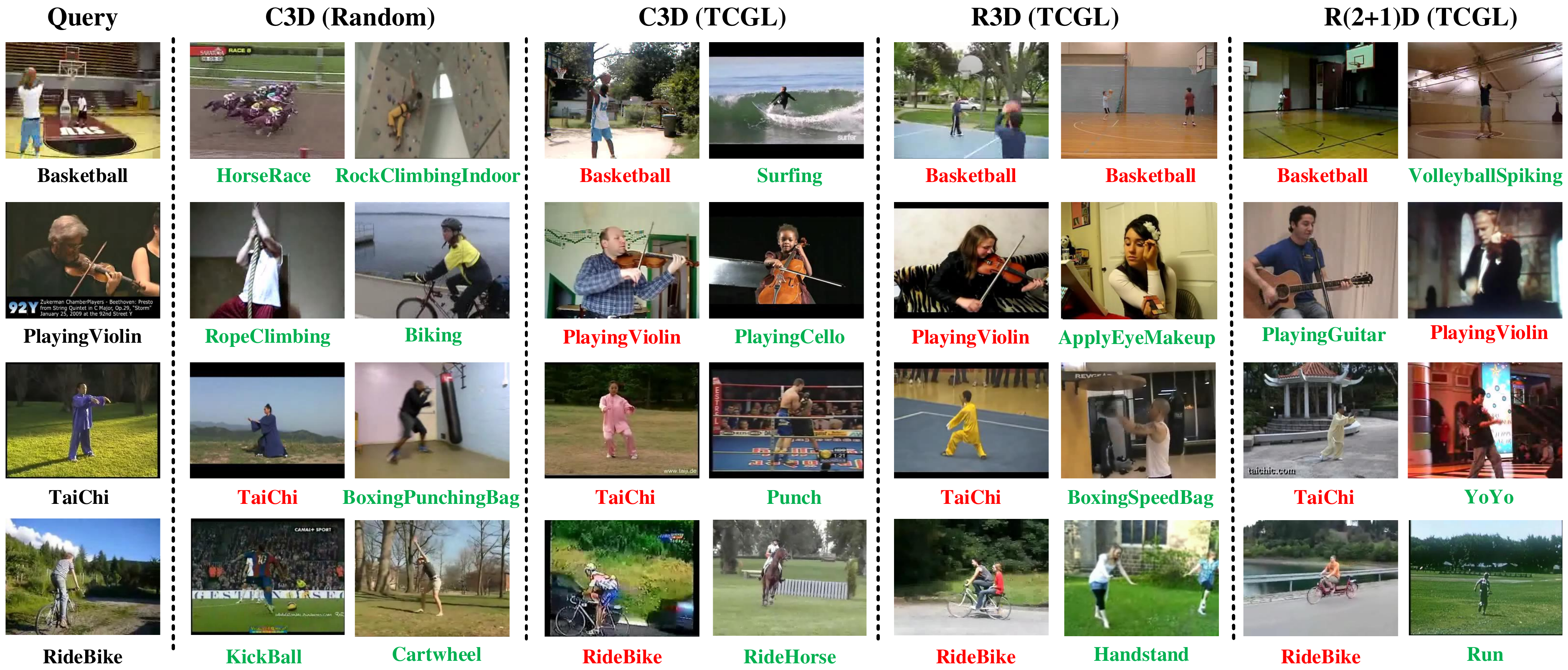}
\end{center}
   \caption{Video retrieval results with TCGL representations on UCF101 and HMDB51 datasets. The first column contains query clips from the test split, and the remaining columns indicate top-2 nearest clips retrieved by different trained models from the training split. The class of each video is displayed in bottom. The class name in red color denotes the correct retrieval class, while the class name in green color denotes the wrong retrieval class. }
\label{Fig7}
\end{figure*}

On Kinetics400 (K400) dataset, our TCGL can achieve performance improvement over MoCo-ImageNet and ImageNet Pre-training. This validates that our TCGL can learn discriminative spatial-temporal representations on Kinetics400 dataset. Since large batch sizes can result in better performances for contrastive learning \cite{yao2020seco}, the $\textrm{CORP}_f$ achieves the best accuracy by using big batchsize (512) and pretraining the model for 800 epochs. The SeCo \cite{yao2020seco} also performs better than our TCGL. This is because that the SeCo also use big batchsize (512) and pretrains the backbone initialized with MoCo on ImageNet. The weights initialized from MoCO on ImageNet can facilitate the pretraining of the SeCo model and thus improve its discriminative power. In addition, the CORP and SeCo need extra data augmentations like random cropping with random scales, color-jitter, random grayscale, blur, and mirror. Different from the CORP and SeCo, our TCGL requires no extra data augmentations and pretrains the backbones from scratch (with random initialized weights), smaller batchsize (16), and fewer pretraining epochs (300). Nonetheless, our TCGL can still achieve comparable performance with $\textrm{CORP}_m$ with similar settings.

On Activity dataset, the SeCo achieves the best performance and performs better than our TCGL due to the fact that the SeCo pretrains the backbone initialized with MoCo on ImageNet and use big batchsize and extra data augmentations. Although pretraining from scratch with small batchsize and no data augmentations, our TCGL can achieve the accuracy of 61.8\% on untrimmed activity recognition dataset ActivityNet, which is more computationally efficient and require less computational resource. These observations validate the advantages of our TCGL in learning discriminative spatial-temporal representations on downstream tasks of action recognition and untrimmed activity recognition.

\begin{table}[!t]\renewcommand\tabcolsep{2pt}\renewcommand\arraystretch{1}
\begin{center}
\begin{tabular}{lccccccc}
\hline
Method&Backbone&Pretrain&Top1&Top5&Top10&Top20&Top50 \\\hline
Jigsaw \cite{noroozi2016unsupervised}&AlexNet&UCF101&19.7&28.5&33.5&40.0&49.4\\
OPN \cite{lee2017unsupervised}&VGG&UCF101&19.9&28.7&34.0&40.6&51.6\\
Deep RL \cite{buchler2018improving}&CaffeNet&UCF101&25.7&36.2&42.2&49.2&59.5\\
SpeedNet \cite{benaim2020speednet}&S3D&K400&13.0&28.1&37.5&49.5&65.0\\\hline
Random &C3D&UCF101&16.7&27.5&33.7&41.4&53.0\\
VCOP \cite{xu2019self}&C3D&UCF101&12.5&29.0&39.0&50.6&66.9\\
PRP \cite{yao2020video}&C3D&UCF101&23.2&38.1&46.0&55.7&68.4\\
V-pace \cite{wang2020self}&C3D&UCF101&20.0&37.4&46.9&58.5&73.1\\
\textbf{TCGL (Ours)}&C3D&UCF101&\textbf{22.4}&\textbf{41.3}&\textbf{51.0}&\textbf{61.4}&\textbf{75.0}\\
\textbf{TCGL (Ours)}&C3D&K400&\textbf{23.6}&\textbf{41.5}&\textbf{52.0}&\textbf{61.8}&\textbf{75.2}\\\hline
Random &R3D-18&UCF101&9.9&18.9&26.0&35.5&51.9\\
VCOP \cite{xu2019self}&R3D-18&UCF101&14.1&30.3&40.4&51.1&66.5\\
PRP \cite{yao2020video}&R3D-18&UCF101&22.8&38.5&46.7&55.2&69.1\\
V-pace \cite{wang2020self}&R3D-18&UCF101&19.9&36.2&46.1&55.6&69.2\\
MemDPC \cite{han2020memory}&R3D-18&UCF101&20.2&40.4&52.4&64.7&-\\
\textbf{TCGL (Ours)}&R3D-18&UCF101&\textbf{23.4}&\textbf{42.2}&\textbf{51.9}&\textbf{62.0}&\textbf{74.8}\\
\textbf{TCGL (Ours)}&R3D-18&K400&\textbf{23.9}&\textbf{43.0}&\textbf{53.0}&\textbf{62.9}&\textbf{75.7}\\\hline
Random &R(2+1)D-18 &UCF101&10.6&20.7&27.4&37.4&53.1\\
VCOP \cite{xu2019self}&R(2+1)D-18 &UCF101&10.7&25.9&35.4&47.3&63.9\\
PRP \cite{yao2020video}&R(2+1)D-18 &UCF101&20.3&34.0&41.9&51.7&64.2\\
V-pace \cite{wang2020self}&R(2+1)D-18 &UCF101&17.9&34.3&44.6&55.5&72.0\\
\textbf{TCGL (Ours)}&R(2+1)D-18 &UCF101&\textbf{21.5}&\textbf{39.3}&\textbf{49.3}&\textbf{59.5}&\textbf{72.7}\\
\textbf{TCGL (Ours)}&R(2+1)D-18 &K400&\textbf{21.9}&\textbf{40.2}&\textbf{49.6}&\textbf{59.7}&\textbf{73.1}\\\hline
\end{tabular}
\end{center}
\caption{Video retrieval result (\%) on UCF101. }
\label{Table12}
\end{table}

\begin{table}[!t]\renewcommand\tabcolsep{2pt}\renewcommand\arraystretch{1}
\begin{center}
\begin{tabular}{lccccccc}
\hline
Method&Backbone&Pretrain&Top1&Top5&Top10&Top20&Top50 \\\hline
Random &C3D&UCF101&7.4&20.5&31.9&44.5&66.3\\
VCOP \cite{xu2019self}&C3D&UCF101&7.4&22.6&34.4&48.5&70.1\\
PRP \cite{yao2020video}&C3D&UCF101&10.5&27.2&40.4&56.2&75.9\\
V-pace \cite{wang2020self}&C3D&UCF101&8.0&25.2&37.8&54.4&77.5\\
MoCo+BE \cite{wang2021removing}&C3D&UCF101&10.2&27.6&40.5&56.2&76.6\\
\textbf{TCGL (Ours)}&C3D&UCF101&\textbf{10.7}&\textbf{28.6}&\textbf{41.1}&\textbf{57.9}&\textbf{77.7}\\
\textbf{TCGL (Ours)}&C3D&K400&\textbf{12.3}&\textbf{30.4}&\textbf{42.9}&\textbf{59.1}&\textbf{79.2}\\\hline
Random &R3D-18&UCF101&6.7&18.3&28.3&43.1&67.9\\
VCOP \cite{xu2019self}&R3D-18&UCF101&7.6&22.9&34.4&48.8&68.9\\
PRP \cite{yao2020video}&R3D-18&UCF101&8.2&25.8&38.5&53.3&75.9\\
V-pace \cite{wang2020self}&R3D-18&UCF101&8.2&24.2&37.3&53.3&74.5\\
MemDPC \cite{han2020memory}&R3D-18&UCF101&7.7&25.7&40.6&57.7&-\\
\textbf{TCGL (Ours)}&R3D-18&UCF101&\textbf{11.7}&\textbf{28.9}&\textbf{40.5}&\textbf{55.4}&\textbf{76.8}\\
\textbf{TCGL (Ours)}&R3D-18&K400&\textbf{12.1}&\textbf{30.6}&\textbf{43.8}&\textbf{58.1}&\textbf{78.0}\\\hline
Random &R(2+1)D-18 &UCF101&4.5&14.8&23.4&38.9&63.0\\
VCOP \cite{xu2019self}&R(2+1)D-18 &UCF101&5.7&19.5&30.7&45.8&67.0\\
PRP \cite{yao2020video}&R(2+1)D-18 &UCF101&8.2&25.3&36.2&51.0&73.0\\
V-pace \cite{wang2020self}&R(2+1)D-18 &UCF101&10.1&24.6&37.6&54.4&77.1\\
\textbf{TCGL (Ours)}&R(2+1)D-18 &UCF101&\textbf{10.5}&\textbf{27.6}&\textbf{39.7}&\textbf{55.6}&\textbf{76.4}\\
\textbf{TCGL (Ours)}&R(2+1)D-18 &K400&\textbf{11.1}&\textbf{30.4}&\textbf{43.0}&\textbf{56.5}&\textbf{77.4}\\\hline
\end{tabular}
\end{center}
\caption{Video retrieval result  (\%) on HMDB51.}
\label{Table13}
\end{table}

\begin{figure*}[t]
\begin{center}
\includegraphics[scale=0.59]{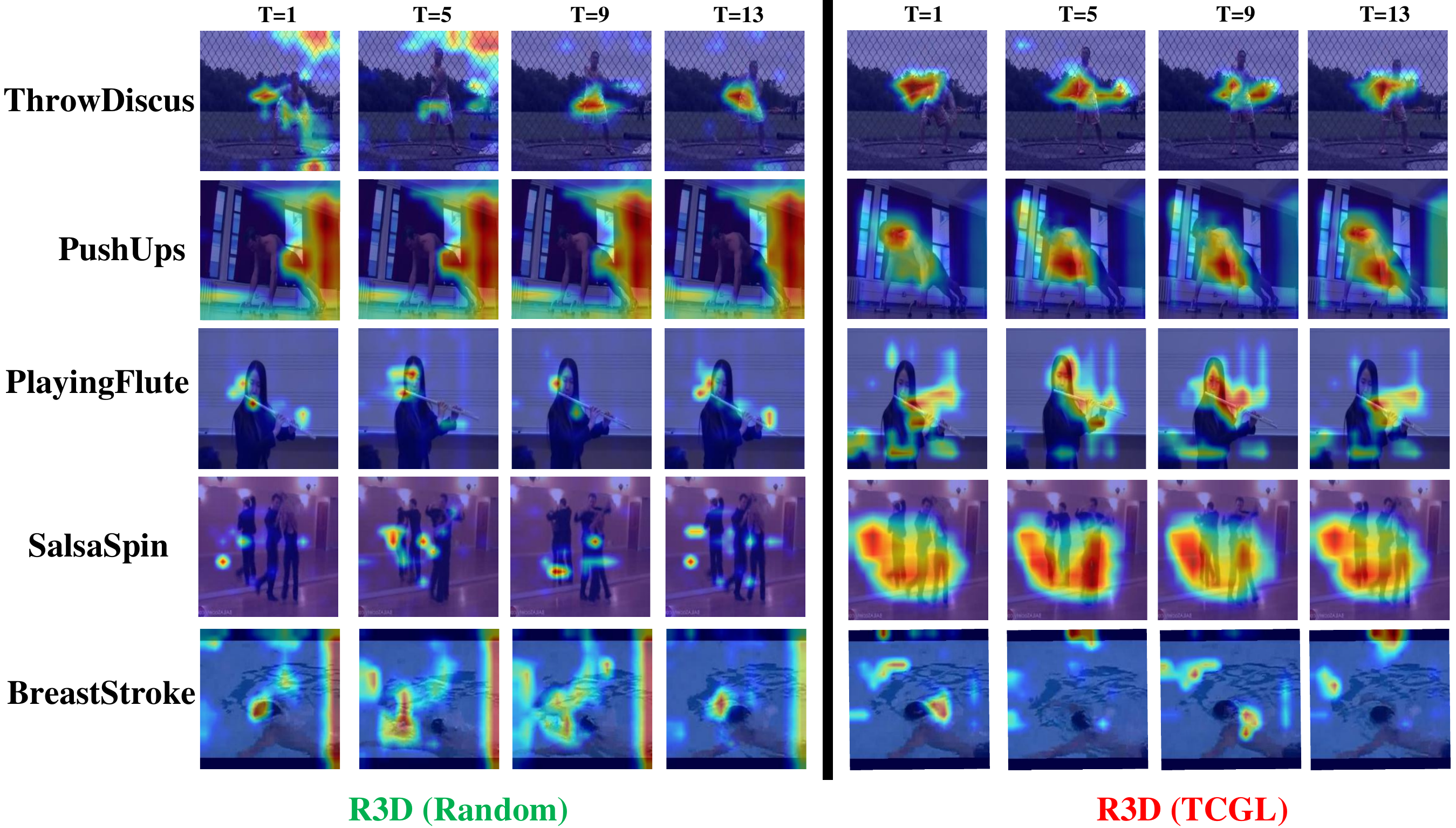}
\end{center}
\vspace{-10pt}
   \caption{Visualization of a sequence of activation maps on UCF101 dataset with R3D-18 as the backbone.}
\label{Fig8}
\end{figure*}

\subsection{Video Retrieval}
To further verify the effectiveness of our TCGL in video retrieval, we test our TCGL on the nearest-neighbor video retrieval. Since the video retrieval task is conducted with features extracted by the backbone network without fine-tuning, its performance largely relies upon the representative capacity of self-supervised models. The experiment is conducted on the first split of UCF101 or the whole training-set of K400, following the protocol in \cite{xu2019self,yao2020video}. In video retrieval, we extract video features from the backbone pre-trained by TCGL. Each video in the testing set is used to query $k$ nearest videos from the training set using the cosine distance. When the class of a test video appears in the classes of $k$ nearest training videos, it is considered as the correct predicted video. We show top-1, top-5, top-10, top-20, and top-50 retrieval accuracies on UCF101 and HMDB51 datasets, and compare our method with other self-supervised methods, as shown in Table \ref{Table12} and \ref{Table13}. For all backbones, our TCGL outperforms the state-of-the-art methods by substantial margins for nearly all metrics. To be noticed, although the SpeedNet \cite{benaim2020speednet} is trained on more powerful backbone S3D, our TCGL performs better. These observations validate our proposed TCGL can learn discriminative representations for video retrieval task.

\subsection{Visualization Analysis}

To have an intuitive understanding of the video retrieval results, Figure \ref{Fig7} visualizes a query video snippet and its top-2 nearest neighbors from the UCF101 dataset. The leftmost columns are videos used for the query, and the remaining columns show top-2 retrieved videos by different feature extractors. In most cases, these retrieved videos portray a strong semantic correlation with the query videos, even when performing complex interactive actions in the presence of significant camera motion, background clutter and occlusion. To be noticed, in some challenging cases, our TCGL fails to retrieve true videos due to the high appearance and semantic similarity. For basketball video, the R(2+1)D network finds volleyball spiking video which is also sports and contains balls. For playing violin video, the C3D network retrieves playing cello video and the R(2+1)D network retrieves playing guitar video, which are also musical activities and contain musical instruments with similar appearance. For taichi video, due to the existence of similar atomic human actions such as crouch, punch, and kick, the C3D, R3D, R(2+1)D networks retrieve punch, boxingspeedbag and yoyo videos, respectively. For ridebike video, due to the similar background color and action movement style, the C3D, R3D, R(2+1)D networks retrieve ridehorse, handshake and run videos, respectively.

To have a better understanding of what TCGL learns, we follow the Class Activation Map \cite{zhou2016learning} to visualize the spatio-temporal regions, as shown in Figure \ref{Fig8}. These examples exhibit a strong correlation between highly activated regions and the dominant movement in the scene. This validates that our TCGL can effectively capture the dominant movement in the videos by learning discriminative temporal representations for videos. To be noticed, in some challenging videos like salsaspin and breaststroke, our TCGL fails to capture true moving regions. In salsaspin videos, our TCGL takes the moving areas in the mirror as the activated regions. In breaststroke videos, our TCGL fails to focus on dominant movement of swimming people due to the disturbance of the motion of the water.

\section{Conclusion}

In this paper, we proposed a novel Temporal Contrastive Graph Learning (TCGL) approach for self-supervised video representation learning. We fully discover discriminative spatial-temporal representation by strict theoretical analysis in the frequency domain based on discrete cosine transform. Integrated with intra-snippet and inter-snippet temporal dependencies, we introduced intra-snippet and inter-snippet temporal contrastive graphs to increase the temporal diversity among video frames and snippets in a graph contrastive self-supervised learning manner. To learn the global context representation and recalibrate the channel-wise features adaptively for each video snippet, we proposed an adaptive video snippet order prediction module, which employs the relational knowledge among video snippets to predict orders. With inter-intra snippet graph contrastive learning strategy and adaptive video snippet order prediction task, the temporal diversity and multi-scale temporal dependency can be well discovered. The proposed TCGL is applied to video action recognition and video retrieval tasks with three kinds of 3D CNNs. Extensive experiments demonstrate the superiority of our TCGL over the state-of-the-art methods on large-scale benchmarks. Future direction will be the evaluation of our method with more powerful backbones, larger pretrained datasets, and more downstream tasks.

\bibliographystyle{IEEEtran}
\bibliography{IEEEabrv,bibfile}

\begin{IEEEbiography}[{\includegraphics[width=1in,height=1.25in,clip,keepaspectratio]{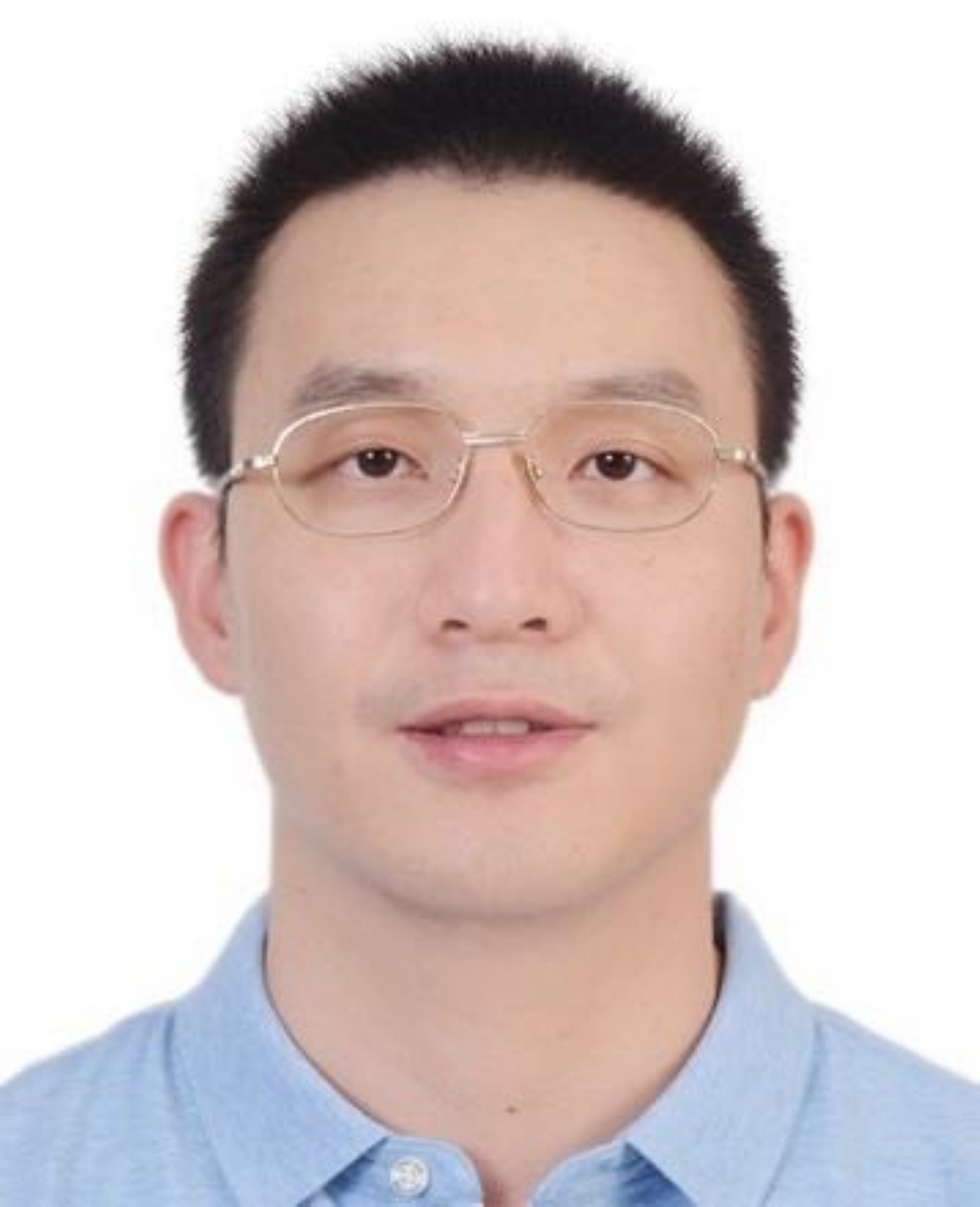}}]{Yang Liu}(M'21) is currently a research associate professor in the School of Computer Science and Engineering, Sun Yat-sen University, working with Prof. Liang Lin. He was a postdoctoral fellow in the School of Computer Science and Engineering, Sun Yat-sen University in 2019-2021. He received the Ph.D. degree in telecommunications and information systems from Xidian University, Xi'an, China in 2019, advised by Prof. Zhaoyang Lu. He received the B.S. degree in telecommunications engineering from Chang'an University, Xi'an, China, in 2014. His current research interests include computer vision and machine learning. He has authored and coauthored more than 10 papers in top-tier academic journals and conferences, including IEEE T-IP, IEEE T-CSVT, and IEEE SPL, etc. More information can be found on https://yangliu9208.github.io/home.
\end{IEEEbiography}

\begin{IEEEbiography}[{\includegraphics[width=1in,height=1.25in,clip,keepaspectratio]{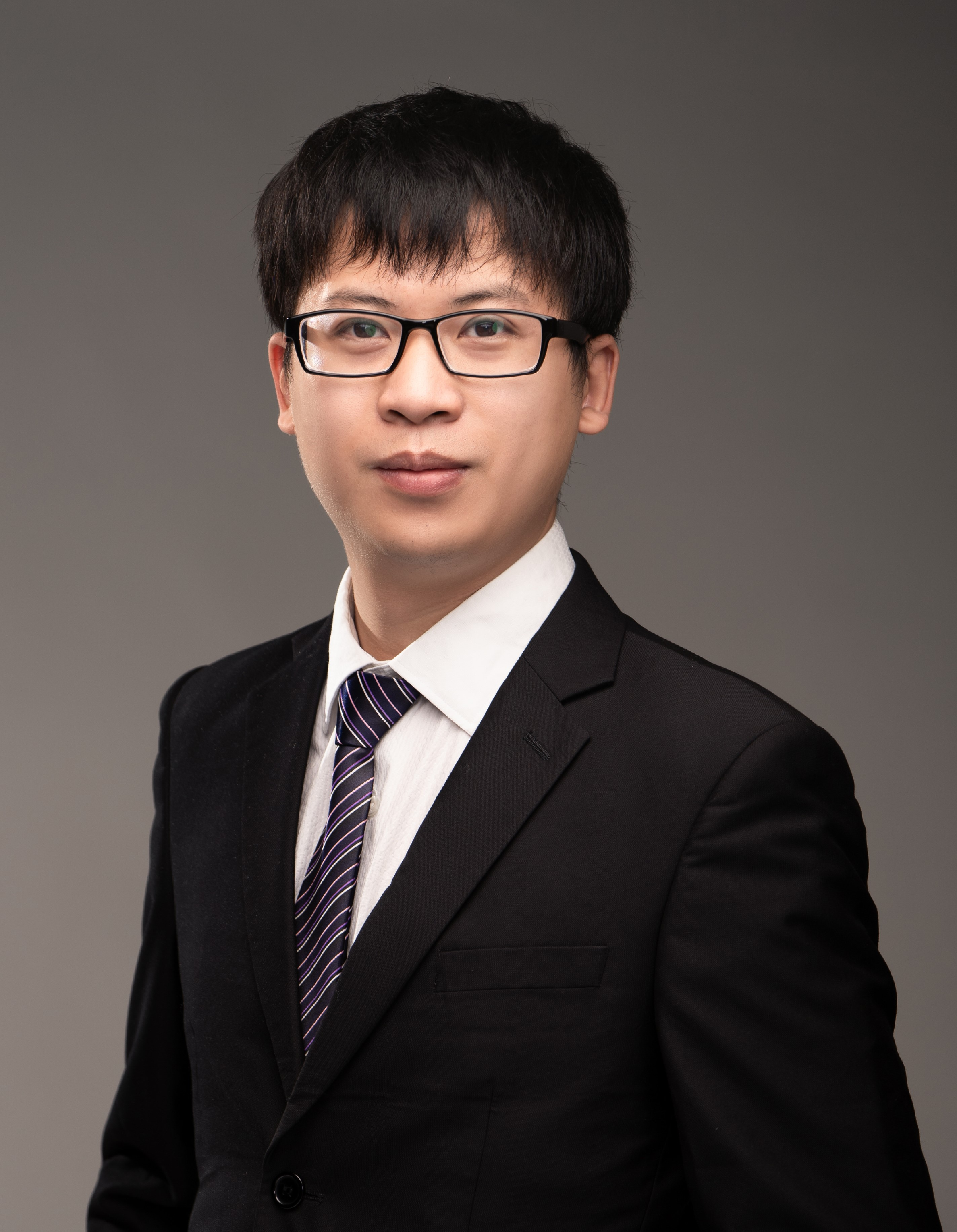}}]{Keze Wang } is currently an associate professor in the School of Computer Science and Engineering, Sun Yat-sen University. He received his B.S. degree in software engineering from Sun Yat-sen University, Guangzhou, China, in 2012. He obtained my Ph.D. degree with honors from the School of Data and Computer Science at Sun Yat-sen University in December 2017, advised by Prof. Liang Lin. He obtained dual PhD awards in the Department of Computing of the Hong Kong Polytechnic University in March 2019, advised by Prof. Lei Zhang. His current research interests include computer vision and machine learning. More information can be found on his personal website https://kezewang.com.
\end{IEEEbiography}

\begin{IEEEbiography}[{\includegraphics[width=1in,height=1.25in,clip,keepaspectratio]{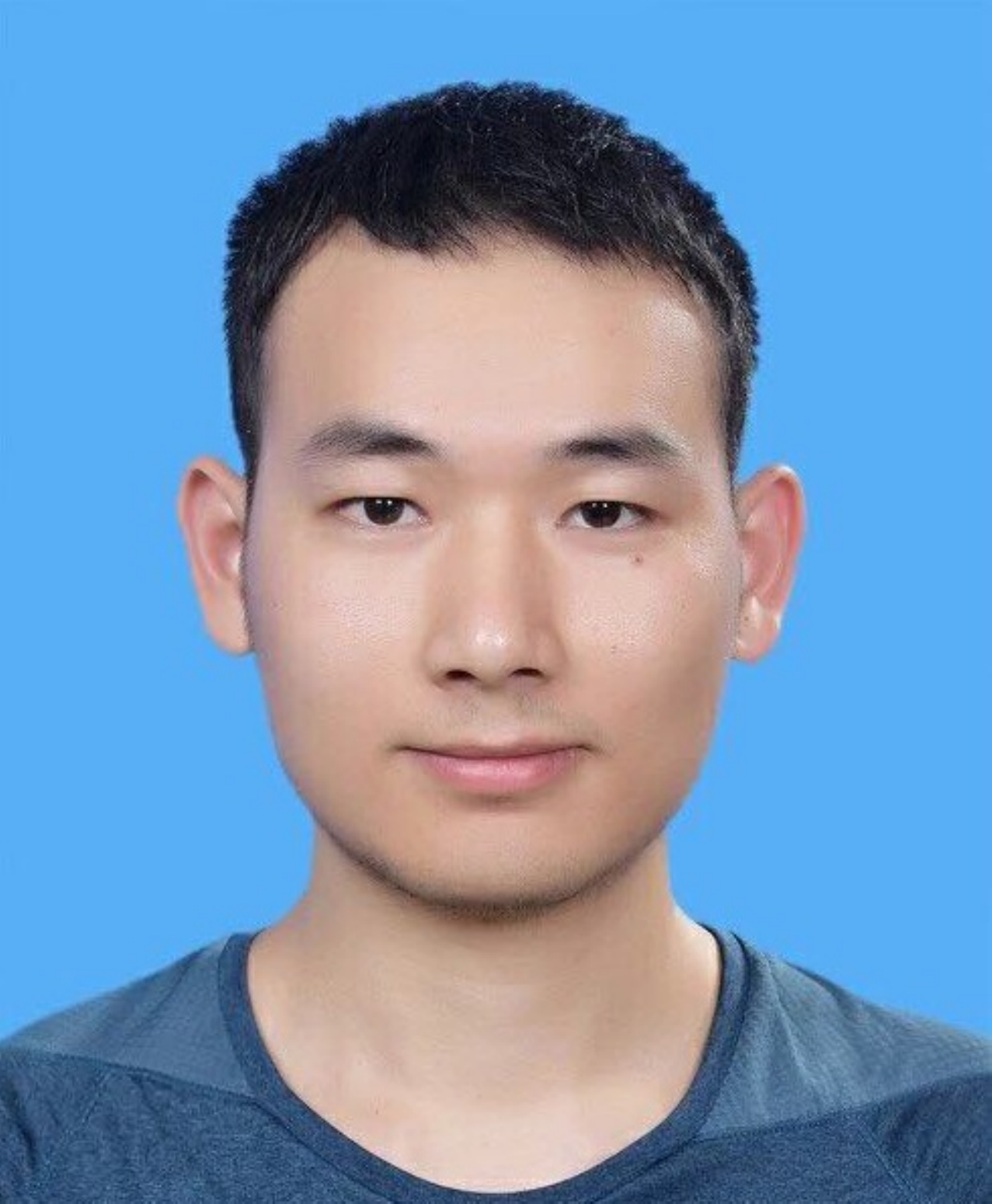}}]{Lingbo Liu} is currently a postdoctoral fellow in the Department of Computing at the Hong Kong Polytechnic University. He received his Ph.D. degree in computer science from Sun Yat-sen University in 2020. He was a research assistant at the University of Sydney, Australia. His current research interests include machine learning and intelligent transportation systems. He has authored and coauthored more than 15 papers in top-tier academic journals and conferences.
\end{IEEEbiography}

\begin{IEEEbiography}[{\includegraphics[width=1in,height=1.25in,clip,keepaspectratio]{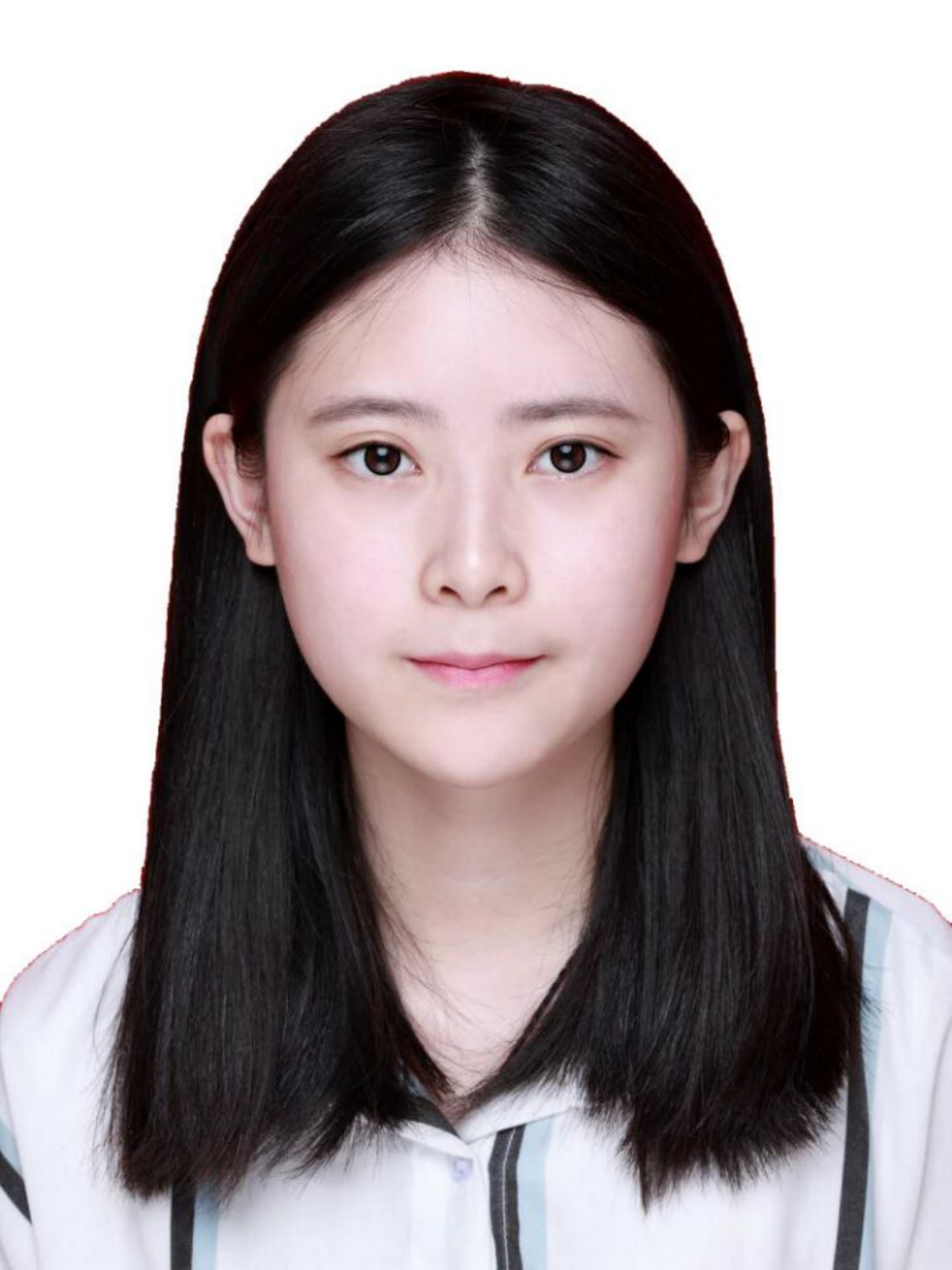}}]{Haoyuan Lan} received her bachelor's degree in Intelligence Science and Technology from Central South University in 2020. She is currently a master student in the School of Computer Science and Engineering, Sun Yat-sen University. Her research interests include computer vision and deep learning.
\end{IEEEbiography}

\begin{IEEEbiography}[{\includegraphics[width=1in,height=1.25in,clip,keepaspectratio]{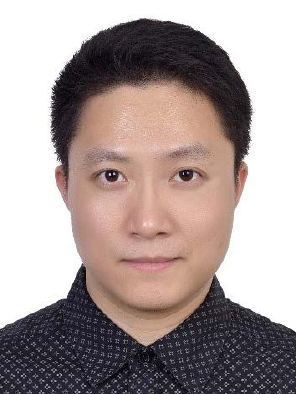}}]{Liang Lin}(M'09, SM'15) is a Full Professor of computer science at Sun Yat-sen University. He served as the Executive Director and Distinguished Scientist of SenseTime Group from 2016 to 2018, leading the R\&D teams for cutting-edge technology transferring. He has authored or co-authored more than 200 papers in leading academic journals and conferences, and his papers have been cited by more than 18,000 times. He is an associate editor of IEEE Trans.Neural Networks and Learning Systems and IEEE Trans. Human-Machine Systems, and served as Area Chairs for numerous conferences such as CVPR, ICCV, SIGKDD and AAAI. He is the recipient of numerous awards and honors including Wu Wen-Jun Artificial Intelligence Award, the First Prize of China Society of Image and Graphics, ICCV Best Paper Nomination in 2019, Annual Best Paper Award by Pattern Recognition (Elsevier) in 2018, Best Paper Dimond Award in IEEE ICME 2017, Google Faculty Award in 2012. His supervised PhD students received ACM China Doctoral Dissertation Award, CCF Best Doctoral Dissertation and CAAI Best Doctoral Dissertation. He is a Fellow of IET.
\end{IEEEbiography}

\end{document}